%% file: main.tex
\documentclass[11pt]{article}

\input{preamble}

\input{macro}

\title{
Logistic Bandits with $\tilde{O}(\sqrt{dT})$ Regret\\
without Context Diversity Assumptions
}

\author{
Seoungbin Bae$^{1}$ \quad Dabeen Lee$^{2}$\\[0.3em]
$^{1}$Department of Industrial \& Systems Engineering, KAIST\\
$^{2}$Department of Mathematical Sciences, Seoul National University\\
\texttt{sbbae31@kaist.ac.kr, dabeenl@snu.ac.kr}
}

\date{}

\begin{document}
\maketitle

\input{sec/introduction}

\input{sec/main_idea}

\input{sec/analysis}
\input{sec/experiments}

\bibliographystyle{plainnat}
\bibliography{ref}

\newpage

\appendix

\input{sec/app_additional_related_work}
\input{sec/app_algorithm2}

\input{sec/app_exp}
\input{sec/app_proofs}

\end{document}

%% file: preamble.tex
\usepackage{fullpage}

\usepackage[T1]{fontenc}
\usepackage[utf8]{inputenc}
\usepackage{lmodern}
\usepackage{microtype}

\usepackage{amsmath,mathtools}
\usepackage{amsfonts,amssymb}
\usepackage{amsthm,thmtools}
\usepackage{bm}
\usepackage{bbm}
\usepackage{accents}

\usepackage{graphicx}
\usepackage{xcolor}
\usepackage{booktabs}
\usepackage{array}
\usepackage{tabularx}
\usepackage{threeparttable}
\usepackage{caption}
\usepackage{subcaption}
\usepackage{makecell}
\usepackage{pifont}

\usepackage{enumitem}

\usepackage{algorithm}
\usepackage{algpseudocode}

\usepackage[round]{natbib}
\usepackage[colorlinks=true,citecolor=blue,linkcolor=red,urlcolor=magenta]{hyperref}
\usepackage{cleveref}

\crefformat{equation}{(#2#1#3)}
\Crefformat{equation}{(#2#1#3)}
\crefrangeformat{equation}{(#3#1#4)--(#5#2#6)}
\Crefrangeformat{equation}{(#3#1#4)--(#5#2#6)}
\crefmultiformat{equation}{(#2#1#3)}{ and (#2#1#3)}{, (#2#1#3)}{, and (#2#1#3)}
\Crefmultiformat{equation}{(#2#1#3)}{ and (#2#1#3)}{, (#2#1#3)}{, and (#2#1#3)}

\theoremstyle{definition}
\newtheorem{theorem}{Theorem}
\newtheorem{lemma}[theorem]{Lemma}
\newtheorem{proposition}[theorem]{Proposition}

\newtheorem{assumption}[theorem]{Assumption}
\newtheorem{remark}[theorem]{Remark}

%% file: macro.tex
\newcommand{\para}[1]{\paragraph{#1.}}

\newcommand{\R}{\mathbb{R}}

\newcommand{\I}{\mathbf{I}}

\newcommand{\st}{\;\mathrel{:}\;}

\DeclarePairedDelimiterX{\cbrp}[1]{\lbrace}{\rbrace}{#1}
\DeclarePairedDelimiterX{\brp}[1]{(}{)}{#1}
\DeclarePairedDelimiterX{\sqbrp}[1]{[}{]}{#1}
\DeclarePairedDelimiterX{\normp}[1]{\lVert}{\rVert}{#1}
\DeclarePairedDelimiterX{\absp}[1]{\lvert}{\rvert}{#1}
\DeclarePairedDelimiterX{\ipp}[1]{\langle}{\rangle}{#1}

\newcommand{\cbr}[1]{\ensuremath{\mathchoice{\cbrp*{#1}}{\cbrp{#1}}{\cbrp{#1}}{\cbrp{#1}}}}
\newcommand{\br}[1]{\ensuremath{\mathchoice{\brp*{#1}}{\brp{#1}}{\brp{#1}}{\brp{#1}}}}
\newcommand{\sqbr}[1]{\ensuremath{\mathchoice{\sqbrp*{#1}}{\sqbrp{#1}}{\sqbrp{#1}}{\sqbrp{#1}}}}
\newcommand{\norm}[1]{\ensuremath{\mathchoice{\normp*{#1}}{\normp{#1}}{\normp{#1}}{\normp{#1}}}}
\newcommand{\abs}[1]{\ensuremath{\mathchoice{\absp*{#1}}{\absp{#1}}{\absp{#1}}{\absp{#1}}}}
\newcommand{\ip}[1]{\ensuremath{\mathchoice{\ipp*{#1}}{\ipp{#1}}{\ipp{#1}}{\ipp{#1}}}}

\DeclareMathOperator*{\argmax}{arg\,max}
\DeclareMathOperator*{\argmin}{arg\,min}

\renewcommand{\P}{\mathbb{P}}
\newcommand{\E}{\mathbb{E}}
\DeclareMathOperator{\Var}{Var}

\DeclareMathOperator{\Bern}{Bern}

\renewcommand{\v}[1]{\boldsymbol{#1}}
\newcommand{\m}[1]{\mathbf{#1}}
\newcommand{\tp}{^{\mathsf{T}}}

\renewcommand{\cal}[1]{\mathcal{#1}}

\newcommand{\what}[1]{\widehat{#1}}
\newcommand{\wtilde}[1]{\widetilde{#1}}

\DeclareRobustCommand{\term}[1]{%
  \ifmmode
    A_{\text{#1}}%
  \else
    term~\ensuremath{A_{\text{#1}}}%
  \fi
}
\newcommand{\uterm}[2]{\ensuremath{\underbrace{#1}_{\term{#2}}}}

\newcommand{\eps}{\epsilon}

\DeclarePairedDelimiterX{\parens}[1]{(}{)}{#1}

\newcommand{\tildeO}[1]{\ensuremath{\mathchoice{\widetilde{\mathcal{O}}\,\parens*{#1}}{\widetilde{\mathcal{O}}\,\parens{#1}}{\widetilde{\mathcal{O}}\,\parens{#1}}{\widetilde{\mathcal{O}}\,\parens{#1}}}}

\algnewcommand\algorithmicinit{\textbf{initialize}}
\algnewcommand\Init{\item[\algorithmicinit]}
\algnewcommand\algorithmicinput{\textbf{input}}
\algnewcommand\Input{\item[\algorithmicinput]}

\newcommand{\note}[1]{{\color{blue}#1}}

%% file: sec/introduction.tex
\begin{abstract}
We study the $K$-armed logistic bandit problem, where at each round, the agent observes $K$ feature vectors associated with $K$ actions. 
Existing approaches that achieve a rate-optimal $\widetilde{\mathcal{O}}(\sqrt{dT})$ regret bound 
rely heavily on context diversity assumptions, such as strict positivity of the minimum eigenvalue of a context covariance matrix.
These assumptions, however, impose strong restrictions on the context process, as they rule out the situation where the context vectors are concentrated in a low-dimensional subspace.
In this paper, we propose SupSplitLog, which, to the best of our knowledge, is the first algorithm for logistic bandits that achieves $\widetilde{\mathcal{O}}(\sqrt{dT})$ regret without any context diversity assumption.
The key idea is to split the collected samples into two disjoint subsets when constructing estimators; one is used to compute an \emph{initial-point estimator}, while the other is used to apply a Newton-type \emph{one-step correction} procedure.
The splitting rule is carefully designed to balance the accuracy requirements of the initial-point estimator and the one-step correction procedure.
Moreover, SupSplitLog strictly improves on the existing algorithms in terms of the dependence on dimension $d$ in the regret upper bound. 
Furthermore, SupSplitLog can be adapted simply to deduce a regret bound that grows with a data-dependent complexity measure, avoiding a direct dependence on $d$, which is favorable when the context vectors are concentrated in a low-dimensional subspace. 
We also provide experimental results that demonstrate numerically the superiority of our algorithm, validating the theoretical results.
\end{abstract}

\section{Introduction}

Logistic bandits provide modeling frameworks for sequential decision-making problems with binary feedback. They naturally arise in many real-world applications and have therefore been extensively studied in recent years \citep{jun2017scalable,liu2024almost,zhang2025generalized,boudart2026enjoying}. Much of the theoretical literature has focused on improving the dependence of regret bounds on problem-dependent quantities, such as $\kappa$ \citep{faury2020improved,abeille2021instance}, where $1/\kappa$ characterizes the minimum variance of the reward distribution, and the norm bound $B$ on the unknown parameter \citep{sawarni2024generalized,lee2024unified}. However, the question of whether we can further reduce the dependence on the context dimension $d$ 
remains relatively underexplored.

Existing approaches that achieve a rate-optimal $\wtilde{\cal O}(\sqrt{dT})$ regret bound for $K$-aremd logistic bandits rely on strong assumptions on the context process \citep{li2017provably,jun2021improved,kim2023double}. They assume that either contexts are i.i.d. or there exists a strictly positive lower bound on the minimum eigenvalue of a context covariance matrix, given by $\lambda_{\min}(\E[\frac{1}{K}\sum_{a\in[K]}\v x_{t,a} \v x_{t,a}\tp])>\sigma_0^2$. These assumptions not only impose strong restrictions on the context process, but also lead to overly conservative regret bounds in high-dimensional problems. In particular, such assumptions forbid the case where context vectors are concentrated around a low-dimensional subspace, in which case the ambient dimension $d$ can be much larger than the dimension of the subspace.


In this paper, we address this gap by developing a new algorithm that guarantees $\wtilde{\cal O}(\sqrt{dT})$ regret without any context diversity assumption. Our algorithm, SupSplitLog, not only lifts the assumption but also attains a regret bound whose dependence on $d$ is strictly better than the existing algorithms. Moreover, SupSplitLog may deal with context vectors concentrated around a low-dimensional subspace, and it can be adapted to deduce a regret bound that scales with a data-dependent complexity measure, avoiding a direct dependence on $d$. SupSplitLog is based on an innovative algorithmic component that splits the incoming samples according to their roles in constructing estimators and applies a Newton-type one-step correction procedure. 
Our contributions are summarized as follows:
\begin{itemize}
    \item We propose SupSplitLog, which is the first algorithm for logistic bandits that achieves a $\wtilde{\cal O}(\sqrt{dT})$ regret bound without any context diversity assumption. Moreover, the regret bound achieves the smallest dependence on $d$ compared to the existing algorithms (see \Cref{table:1}).
    
    \item 
    Moreover, the algorithm can be adapted to deduce a regret bound that grows with a data-dependent complexity measure such as $\log \frac{\det \m V_T}{\det \kappa \lambda \I}$, where $\m V_T$ denotes the Gram matrix formed by the realized feature vectors, thereby avoiding a direct dependence on $d$. 

    \item The main technical idea is to split the collected samples into two disjoint subsets when constructing estimators: one subset is used to compute an \emph{initial-point estimator}, while the other is used to perform a Newton-type \emph{one-step correction} procedure. The splitting rule is carefully designed to balance the accuracy requirements between the initial-point estimator and the one-step correction framework. This novel technique of one-step correction leads to a high-probability bound on the estimation error that does not depend on $d$.

    \item We complement our theoretical analysis with experiments on high-dimensional problems with low-dimensional structure, where the realized complexity $\log \frac{\det \m V_T}{\det \kappa \lambda \I}$ is small. The results show that SupSplitLog performs most favorably among the baselines in such regimes, consistent with our theory.
\end{itemize}

\begin{table}[t]
\centering
\caption{Comparison with prior work. ``Data-dependent'' indicates whether the regret bound admits a refinement in terms of a data-dependent complexity measure, such as $\log \frac{\det \m V_T}{\det\kappa \lambda \I}$, where $\m V_T$ denotes the Gram matrix formed by the realized feature vectors up to round $T$.}
\label{table:1}
\small
\begin{tabular}{l c c c}
\toprule
Algorithm
& Diversity assumption ({$\sigma_0^{-2} = \Omega(d)$})
& {Regret $\widetilde{\mathcal O}(\cdot)$}
& Data-dependent \\
\midrule
\makecell[l]{SupCB-GLM\\ \citep{li2017provably}}
& {$\lambda_{\min}(\E[\frac{1}{K}\sum_{a\in[K]}\v x_{t,a} \v x_{t,a}\tp])>\sigma_0^2$}
& {$\kappa\sqrt{dT}+d \sigma_0^{-8}+\kappa^8 d^3 \sigma_0^{-4}$}
& \ding{55} \\

\makecell[l]{SupLogistic\\\citep{jun2021improved}}
& {$\lambda_{\min}(\E[\frac{1}{K}\sum_{a\in[K]}\v x_{t,a} \v x_{t,a}\tp])>\sigma_0^2$}
& {$\sqrt{dT}+d \sigma_0^{-8}+\kappa^2 d \sigma_0^{-4}$}
& \ding{55} \\

\makecell[l]{DDRTS-GLM\\\citep{kim2023double}}
& {$\lambda_{\min}(\E[\frac{1}{K}\sum_{a\in[K]}\v x_{t,a} \v x_{t,a}\tp])>\sigma_0^2$}
& {$\sqrt{\kappa\sigma_0^{-2}T}+\kappa^3 \sigma_0^{-6}$}
& \ding{55} \\

\makecell[l]{SupSplitLog\\(ours)}
& -
& {$\kappa\sqrt{dT}+\kappa^{2.5} d^{2.5}$}
& \ding{51} \\
\bottomrule
\end{tabular}
\end{table}

\section{Related Work} \label{sec:related_work}


Although there is a regret lower bound of $\Omega(d\sqrt{T})$ for the general (infinite-armed) setting of logistic bandits~\citep{abeille2021instance}, the $K$-armed setting admits sharper bounds of order $\wtilde{\cal O}(\sqrt{dT})$ \citep{li2017provably,jun2021improved}, and these match a lower bound of order $\Omega(\sqrt{dT})$ for $K$-armed generalized linear bandits~\citep{li2019nearly}.
This improvement is made possible by what we call the Auer framework \citep{auer2002using}, which constructs independent sample buckets and thereby allows one to apply scalar-valued concentration inequalities for independent noises, instead of the vector-valued concentration arguments typically used in the general setting. Building on the Auer framework, \citet{li2017provably} first established a regret bound of order $\wtilde{\cal O}(\kappa \sqrt{dT} + \kappa^8 d^3 \sigma_0^{-4} + d \sigma_0^{-8})$ for $K$-armed logistic bandits. Later, \citet{jun2021improved} improved the dependence on the problem-dependent quantity $\kappa$, where $1/\kappa$ characterizes the minimum variance of the reward distribution (see \Cref{ass:kappa} for a formal definition), and obtained a regret bound of order $\wtilde{\cal O}(\sqrt{dT} + \kappa d\sigma_0^{-4} + d\sigma_0^{-8})$. However, both results rely on the context diversity assumption $\lambda_{\min}\br{\E\sqbr{\frac{1}{K}\sum_{a\in[K]}\v x_{t,a} \v x_{t,a}\tp}}>\sigma_0^2$, which becomes problematic in high-dimensional regimes since $\Omega(\sigma_0^{-2})=d$.

More recently, \citet{kim2023double} proposed a different approach based on doubly robust estimation and Thompson sampling, taking a detour from the Auer framework, and their regret bound is $\wtilde{\cal O}(\sigma_0^{-1}\sqrt{\kappa T} + \kappa^3 \sigma_0^{-6})$. Nevertheless, their analysis still requires the context diversity assumption, and $\sigma_0^{-1}\sqrt{T}$ can be worse than $\sqrt{dT}$ as $\Omega(\sigma_0^{-2})=d$. 

There exist algorithms for generalized linear bandits and infinite-armed logistic bandits that do not rely on the context diversity assumption, but their regret bounds are of order $\wtilde{\cal O}(d\sqrt{T})$~\citep{filippi2010parametric}.
As in \Cref{table:1}, our algorithm achieves a regret bound of order $\wtilde{\cal O}(\kappa\sqrt{dT} + \kappa^{2.5}d^{2.5})$ without the context diversity assumption. Moreover, as $\Omega(\sigma_0^{-2})=d$, our regret bound improves on the previous bounds in terms of dependence on $d$. We list additional related work in \Cref{sec:additional_related_work}.

\section{Problem Setting}

We consider the $K$-armed contextual bandit problem where for each round $t\in[T]$, the agent observes an arm set $\cal X_t\coloneqq\{\v x_{t,a}\}_{a\in[K]}\subset\R^d$, chooses $a_t\in[K]$, and receives a stochastic reward $r_t\in\{0,1\}$ given by $r_{t}\mid \cal F_t, \v x_{t,a_t} \sim \Bern(\mu(\v x_{t,a_t}\tp\v\theta^*))$. Here, $\cal F_t=\sigma(\{\cal X_i,a_i\}_{i=1}^t,\{r_i\}_{i=1}^{t-1})$ defines a filtration where $\v x_{t,a_t}$ is $\cal F_t$-measurable and $r_t$ is $\cal F_{t+1}$-measurable. In addition, $\mu(z) = (1+e^{-z})^{-1}$ is the logistic function, and $\v\theta^*\in\R^d$ is an unknown parameter.
Let $a_t^*$ be an optimal arm in round $t$, i.e., $a_t^* = \argmax_{a\in[K]} \v x_{t,a}\tp\v\theta^*$. The agent's goal is to minimize cumulative regret, defined as $R_T \coloneqq \sum_{t=1}^T [\mu(\v x_{t,a_t^*}\tp\v\theta^*)-\mu(\v x_{t,a_t}\tp\v\theta^*)]$. 

The following are some standard assumptions used throughout the paper. 

\begin{assumption} \label{ass:bound}
    $\|\v x\|_2\leq 1$ for all $\v x\in\cal X_t, t\in[T]$, and $\v\theta^* \in \Theta\coloneqq\{\v\theta \st \|\v \theta\|_2 \leq B\}$ for some $B>0$. Here, $\v \theta^*$ is unknown  while $B$ is known to the agent.
\end{assumption}

\begin{assumption} \label{ass:kappa}
For some $\kappa,L_\mu>0$, we have $1/\kappa \leq \dot\mu(\v x\tp \v\theta) \leq L_{\mu}$ for all $\v x\in\cal X_t, t\in[T]$, $\v\theta\in\Theta$.
\end{assumption}

\Cref{ass:bound} impose bounds on the size of feature vectors and the unknown parameter $\v \theta^*$. \Cref{ass:kappa} introduces instance-dependent constants that characterize the local behavior of $\dot\mu(\cdot)$. Here, we set $L_\mu=1/4$ to simplify the proof for the base case of \Cref{lem:per-round}.

\begin{assumption} \label{ass:dist}
    We assume that the context sequence is generated independently of the agent's past actions and realized rewards.
\end{assumption}

\begin{remark}[Remark on \Cref{ass:dist}]
Previous works \citep{li2017provably,jun2021improved,kim2023double} assume that (i) $\cal X_1,\dots,\cal X_T$ are i.i.d. with a fixed distribution $\nu$, and (ii) the minimum eigenvalue of the context covariance matrix is uniformly lower bounded by a strictly positive quantity, i.e., $\lambda_{\min}(\E[\frac{1}{K}\sum_{a\in[K]}\v x_{t,a} \v x_{t,a}\tp])>\sigma_0^2$ with $\sigma_0^{-2}=\Omega(d)$. We impose none of these assumptions. \Cref{ass:dist} is strictly weaker, and in fact, the context diversity assumption automatically implies it.
\end{remark}

%% file: sec/main_idea.tex
\section{Challenges in Achieving $\wtilde O(\sqrt{dT})$ Regret in Logistic Bandits} \label{sec:challenges}

In this section, we identify the key requirement for achieving $\wtilde O(\sqrt{dT})$ regret in logistic bandits, explain the main technical obstacle in establishing this requirement, and review the approach of previous work together with its limitations. This will help us to better motivate our framework.

To achieve $\wtilde O(\sqrt{dT})$ regret, we need a high-probability upper bound on the error term $|\v x\tp (\what{\v\theta}_t - \v\theta^*)|$ that does not depend on $d$ for a fixed $\v x\in\R^d$ and a given estimator $\what{\v\theta}_t$ after round $t$~\citep{li2017provably,jun2021improved}. 
When $\what{\v\theta}_t$ is the maximum likelihood estimator of the logistic loss $\cal L_t(\v\theta)=-\sum_{i=1}^t [r_i \log(\mu(\v x_{i,a_i}\tp\v\theta)) + (1-r_i) \log(1-\mu(\v x_{i,a_i}\tp\v\theta))]$, we may argue that $\abs{\v x\tp \br{\what{\v\theta}_t - \v\theta^*}} =  \abs{\sum_{i=1}^t   z_i \eps_i}$ 
where $z_i=\br{\v x\tp \check {\m H}_t^{-1}\v x_{i,a_i}}$, $\check{\m H}_t = \sum_{i=1}^t \dot\mu(\v x_{i,a_i}\tp \check{\v\theta}_{t,i}) \v x_{i,a_i} \v x_{i,a_i}\tp$ is an estimator-induced Hessian matrix, $\check{\v\theta}_{t,i} = (1-c_i) \what{\v\theta}_t+c_i \v\theta^*$ for some $c_i\in[0,1]$ due to the mean-value theorem, and $\eps_i=r_i - \mu(\v x_{i,a_i}\tp\v\theta^*)$. 
Then one may attempt to apply a scalar-valued concentration inequality to $|\sum_{i=1}^t z_i \eps_i|$ to yield a high-probability upper bound that does not depend on $d$. 
However, the difficulty is two-fold. First, under a general bandit policy, the realized rewards $r_1,\dots,r_i$ affect future selected actions and observed rewards. 
Second, even if the policy is independent of the realized rewards, the definition of $\check{\m H}_t$ shows that $z_i$ still depends on future intermediate points $\check{\v\theta}_{t,i+1}, \dots, \check{\v\theta}_{t,t}$. 
Therefore, scalar-valued concentration inequalities cannot be applied directly.

To circumvent this issue, the existing algorithms \citep{li2017provably,jun2021improved} replaced $\check{\m H}_t$ with the true Hessian $\m H_t^* = \sum_{i=1}^t \dot\mu(\v x_{i,a_i}\tp\v\theta^*) \v x_{i,a_i} \v x_{i,a_i}\tp$, leading to
\begin{align}
    \abs{\v x\tp \br{\what{\v\theta}_t - \v\theta^*}} \leq  \underbrace{\abs{\sum_{i=1}^t   \br{\v x\tp \m H_t^{*-1} \v x_{i,a_i}} \cdot \eps_i}}_{\mathrm{(noise)}} + \underbrace{\abs{\v x\tp \m H_t^{-1} \br{\check{\m H}_t - \m H_t^*} \br{\v\theta^* - \what{\v\theta}_t} }}_{\mathrm{(remainder)}}. \label{eq:diff1}
\end{align}
For the (noise) term, they adapted the Auer framework \citep{auer2002using} to construct level-wise buckets so that samples within each bucket are selected independently of the realized rewards. One can then apply a scalar-valued concentration inequality within each bucket. For the (remainder) term, the key is to keep $\check{\m H}_t$ and $\m H_t^*$ close to each other. Since each $\check{\v\theta}_{t,i}$ lies between $\what{\v\theta}_t$ and $\v\theta^*$, it is enough to keep $\max_{i\in[t]} \abs{\v x_{i,a_i}\tp(\what{\v\theta}_t - \v\theta^*)}$ small~\citep{jun2021improved}.
Applying the Cauchy-Schwarz inequality, we have
\begin{align}
    \max_{i\in[t]} \abs{\v x_{i,a_i}\tp(\what{\v\theta}_t - \v\theta^*)} \leq \max_{i\in[t]} \|\v x_{i,a_i} \|_{\m V_{t}^{-1}} \cdot \|\what{\v\theta}_t - \v\theta^* \|_{\m V_{t}} \leq  \lambda^{-1/2}_{\min}(\m V_t) \cdot \|\what{\v\theta}_t - \v\theta^* \|_{\m V_{t}} \label{eq:diff2}
\end{align}
where $\m V_t = \sum_{i=1}^t \v x_{i,a_i} \v x_{i,a_i}\tp$. Here, to ensure that $\lambda_{\min}(\m V_t)$ is sufficiently large, the existing methods apply an initial random-sampling phase whose length grows as a function of $d$.

However, the above approach relies on the assumption that $\lambda_{\min}(\E[\frac{1}{K}\sum_{a=1}^K \v x_{i,a} \v x_{i,a}\tp])\geq \sigma_0^2$ and cannot avoid additional random-sampling rounds. These requirements are restrictive in two respects. First, they exclude settings where the features are 
concentrated around a lower-dimensional subspace. Second, the random-sampling phase is statistically inefficient and can be impractical.
More fundamentally, both ingredients enforce uniform exploration across all $d$ dimensions of the feature space.
Consequently, the regret guarantees scale with the ambient dimension $d$ and can be overly conservative in high-dimensional settings.
To overcome these limitations, in the next section we propose a new algorithm that separates the roles of the collected samples, using one part to obtain an initial estimator and the other to carry out a one-step correction, thereby avoiding the context diversity assumption and a random-sampling phase.

\section{Main Idea and Algorithm Design}

\begin{algorithm}[t]
\caption{SupSplitLog}
\label{alg:1}
\begin{algorithmic}[1]
\Input $d,T,K,B,\kappa,\lambda,\delta$
\Init $S\gets\lfloor 0.5 \log_2T\rfloor$, $\alpha$ as in \Cref{eq:alpha}, $\beta$ as in \Cref{eq:beta}
\Init For $s\in[S]$,~~~~set $\tau^{(s)}$ as in \Cref{eq:tau}, $\Psi^{(s)}\gets\emptyset$, $P^{(s)}\gets\emptyset$, $E^{(s)}\gets\emptyset$, $\bar{\v\theta}^{(s)}_0\gets \v 0$
\For{$t=1,\dots,T$}
  \State Initialize $A^{(1)}\gets[K]$, $s\gets 1$, $a_t\gets \mathtt{None}$
  \While{$a_t=\mathtt{None}$}
    \State $\m V^{(s)}_{P,t-1} \gets \kappa \lambda \I + \sum_{i\in P^{(s)}(t-1)} \v x_{i,a_i} \v x_{i, a_i}\tp$
    \State $\m V^{(s)}_{E,t-1} \gets \kappa \lambda \I + \sum_{i\in E^{(s)}(t-1)} \v x_{i,a_i} \v x_{i, a_i}\tp$
    \State Update $\bar{\v\theta}_{t-1}^{(s)}$ as in \Cref{eq:theta_bar},~~~~$\what{\v\theta}_{t-1}^{(s)}$ as in \Cref{eq:theta_hat}
    \State For $a\in A^{(s)}$,~~~~compute $m^{(s)}_{t,a}\gets \v x_{t,a}\tp \what{\v\theta}_{t-1}^{(s)}$,~~~~$w^{(s)}_{t,a}\gets \alpha \|\v x_{t,a}\|_{(\m V_{E,t-1}^{(s)})^{-1}}$
    \If{$\exists a\in A^{(s)}\st w^{(s)}_{t,a}>2^{-s}$} \Comment{condition (a)}
      \State Choose $a_t\gets \argmax_{a\in A^{(s)}} w^{(s)}_{t,a}$,~~~~receive $r_t$
      \State $\Psi^{(s)}\leftarrow \Psi^{(s)} \cup \{t\}$
      \If {$\|\v x_{t,a_t}\|_{(\m V_{P,t-1}^{(s)})^{-1}}^2 > \tau^{(s)}$} \Comment{condition (a1)}
        \State $P^{(s)}\leftarrow P^{(s)} \cup \{t\}$
      \Else \Comment{condition (a2)}
        \State $E^{(s)}\leftarrow E^{(s)} \cup \{t\}$
      \EndIf
    \ElsIf{$\forall a\in A^{(s)}~,~ w^{(s)}_{t,a}\le 1/\sqrt{T}$} \Comment{condition (b)}
      \State Choose, $a_t\gets\argmax_{a\in A^{(s)}} m^{(s)}_{t,a}$,~~~~receive $r_t$
    \Else \Comment{condition (c)}
      \State $A^{(s+1)}\gets\{a\in A^{(s)} \st m^{(s)}_{t,a}\geq \max_{a\in A^{(s)}}m^{(s)}_{t,a}-2\cdot 2^{-s}\}$
      \State $s\gets s+1$
    \EndIf
  \EndWhile
\EndFor
\end{algorithmic}
\end{algorithm}

In this section, we introduce our algorithm, SupSplitLog, in \Cref{alg:1}. As in the Auer framework, the algorithm maintains, for each level $s\in[S]$, a bucket $\Psi^{(s)}$ that stores the time indices of the arms selected at that level. However, the key distinction from the existing sup-style algorithms is that we partition the samples of each bucket into two disjoint subsets, namely the pilot set $P^{(s)}$ and the estimation set $E^{(s)}$, i.e., $\Psi^{(s)} = P^{(s)}\cup E^{(s)}$ and $P^{(s)}\cap E^{(s)} = \emptyset$. These two subsets are used to construct an initial-point estimator $\bar{\v\theta}^{(s)}_{t-1}$ and a one-step corrected estimator $\hat{\v\theta}^{(s)}_{t-1}$, respectively. 
A newly collected sample is assigned to $P^{(s)}$ or $E^{(s)}$ according to a level-dependent threshold $\tau^{(s)}$, which balances the need to improve the initial-point estimator against the need to perform an accurate one-step correction.
This sample-splitting design is the key structural ingredient behind our analysis.

We set $\lambda=1$. At level $s$ in round $t$, we first construct the design matrices $\m V_{P,t-1}^{(s)}=\kappa \lambda \I + \sum_{i\in P^{(s)}(t-1)} \v x_{i,a_i} \v x_{i,a_i}\tp$ and  $\m V_{E,t-1}^{(s)}=\kappa \lambda \I + \sum_{i\in E^{(s)}(t-1)} \v x_{i,a_i} \v x_{i,a_i}\tp$, and then we use the samples in $P^{(s)}(t-1)$ to compute the \emph{initial-point estimator} $\bar{\v\theta}^{(s)}_{t-1}$ as follows:
\begin{align}
    \bar{\v\theta}_{t}^{(s)} = \argmin_{\v\theta\in\Theta} \cbr{\sum_{i\in P^{(s)}(t)} \sqbr{- r_i \log\br{\mu(\v x_{i,a_i}\tp\v\theta)} - (1-r_i) \log \br{1 - \mu(\v x_{i,a_i}\tp \v\theta)}} + \frac{\lambda}{2} \|\v\theta\|_2^2 }. \label{eq:theta_bar}
\end{align}
We then update $\bar{\v\theta}^{(s)}_{t-1}$ using the samples in $E^{(s)}(t-1)$ through a \emph{one-step correction}, which yields
\begin{align}
    \what{\v\theta}_{t}^{(s)} = \bar{\v\theta}_t^{(s)} + \br{\m H_t^{(s)} (\bar{\v\theta}_t^{(s)})}^{-1} \v g_t^{(s)}(\bar{\v\theta}_t^{(s)}), \label{eq:theta_hat}
\end{align}
where the gradient $\v g_t^{(s)}(\v\theta)$ and the Hessian $ \m H_t^{(s)} (\v\theta)$ of the loss function are given by
\begin{align*}
    \quad \v g_t^{(s)}(\v\theta) =\sum_{i\in E^{(s)}(t)} \br{r_i - \mu(\v x_{i,a_i}\tp \v\theta)}\v x_{i,a_i} - \lambda \v\theta, \quad \m H_t^{(s)} (\v\theta) = \lambda \I + \sum_{i\in E^{(s)}(t)} \dot\mu(\v x_{i,a_i}\tp \v\theta) \v x_{i,a_i} \v x_{i,a_i}\tp.
\end{align*}

\begin{remark}[Discussion on the one-step estimator $\what{\v\theta}$]
A one-step estimator starts from an initial estimator and applies a single Newton-type correction step, using the gradient and curvature information, instead of fully re-optimizing the objective. This idea is classical in asymptotic and semiparametric statistics, where a single correction is often enough to recover much of the accuracy of the exact estimator \citep{bickel1993efficient,van2000asymptotic}. 
We adapt this idea in a bandit-specific way through sample splitting. The initial-point estimator $\bar{\v\theta}_t^{(s)}$ is computed from the pilot set, while the one-step correction $\what{\v\theta}_t^{(s)}$ is carried out on the estimation set. As a result, the matrix quantity $\m H_t^{(s)}(\bar{\v\theta}_t^{(s)})$ is determined by the pilot estimator and the observed contexts, and it is separated from the realized rewards in the estimation set. This separation is the key to our concentration argument.
\end{remark}

\begin{remark}[Comment on regret analysis] An additional residual term appears in regret analysis (see the (remainder) term in \Cref{lem:decomp}), due to the correction step applied to the exact estimator for the estimation set. Our splitting rule, through conditions (a1) and (a2), is designed to control this, by ensuring that the pilot samples make the initial-point estimator sufficiently accurate before the correction step. Thus, the same split simultaneously enables concentration of the main stochastic term and keeps the residual term small. We explain more details in \Cref{sec:analysis}. 
\end{remark}

For each arm $a\in A^{(s)}$, we compute $m_{t,a}^{(s)}=\v x_{t,a}\tp \what{\v\theta}_{t-1}^{(s)}$ and $w_{t,a}^{(s)}=\alpha \| \v x_{t,a}\|_{(\m V_{E,t-1}^{(s)})^{-1}}$, where
\begin{align}
    \alpha = 2(\alpha_1 + \alpha_2),~~ \alpha_1 &= \sqrt{2L_\mu \kappa^2 \log(4TSK/\delta)}  + \sqrt{\kappa/(9\lambda)}\log(4TSK/\delta), ~~ \alpha_2 = B \sqrt{\kappa\lambda} \label{eq:alpha}. 
\end{align}
The confidence level associated with level $s$ is set to $2^{-s}$. The algorithm then proceeds as follows: [condition (a)] If the uncertainty with respect to $E^{(s)}$ is still large in the direction of $\v x_{t,a}$, i.e., $w_{t,a}^{(s)} > 2^{-s}$, then the algorithm explores and adds the current time index to the bucket, $\Psi^{(s)}\gets \Psi^{(s)}\cup\{t\}$. [condition (b)] If condition (a) does not hold and the uncertainty of every action is below $1/\sqrt{T}$, then we stop and proceed to exploitation. [condition (c)] Otherwise, if the uncertainty of every action is below the current confidence level, then the algorithm eliminates arms that cannot be optimal, namely those satisfying $m_{t,a}^{(s)} < \max_{a\in A^{(s)}} m _{t,a}^{(s)} - 2 \cdot 2^{-s}$, and moves to the next level, i.e., $s\gets s+1$.

When condition (a) is triggered, we further decide whether the newly collected sample should be assigned to the pilot set or the estimation set. [condition (a1)] If the uncertainty with respect to $P^{(s)}$ in the direction $\v x_{t,a_t}$ exceeds the level-dependent threshold $\tau^{(s)}$, i.e. $\|\v x_{t,a_t}\|_{(\m V_{P,t-1}^{(s)})^{-1}}^2 > \tau^{(s)}$,
\begin{align}
     \tau^{(s)} &= \frac{1}{\beta^2} \min\cbr{1,~ \frac{2^{-s}}{\sqrt{32 L_{\mu} \kappa d \log(1+T/(\kappa \lambda d))}}}, \label{eq:tau}  \\
     \beta &= \kappa\sqrt{0.5d\log(1+T/(\kappa\lambda d)) +  0.5\log (2S/\delta)} + B\sqrt{\kappa \lambda}, \label{eq:beta}
\end{align}
then we add the corresponding time index to $P^{(s)}$ to improve the accuracy of the initial-point estimator. [condition (a2)] Otherwise, we add it to $E^{(s)}$ so that it can be used for the one-step correction.

We next state the main regret bound for SupSplitLog. The full proof is deferred to \Cref{sec:regret_analysis}.

\begin{theorem} \label{thm:regret1}
    With probability at least $1-\delta$, the regret of \Cref{alg:1} is upper bounded by
    \begin{align*}
        R_T &\leq 16 L_\mu  d \log (1+T/(\kappa \lambda d)) \beta^2 \br{1 + S\sqrt{32L_\mu \kappa d \log(1+T/(\kappa \lambda d))}} \\
        &\quad + 8L_\mu \alpha \sqrt{2 S T d \log (1 + T/(\kappa \lambda d))} + 2L_\mu \sqrt{T}.
    \end{align*}
\end{theorem}

\begin{remark} [Data-dependent regret of \Cref{thm:regret1}] \label{remark:data-dependent}
While \Cref{thm:regret1} is stated in terms of the ambient dimension $d$, this dependence can be refined. By allowing $\beta$ and $\tau^{(s)}$ to be chosen adaptively based on the data observed up to round $t$, e.g., through $\beta_t^{(s)}$ and $\tau_t^{(s)}$, one can refine the bound so that $d$ is replaced by a data-dependent complexity measure of $\log \frac{\det \m V_T}{\det \kappa \lambda \I}$ where $\m V_T= \sum_{t=1}^T \v x_{t,a_t} \v x_{t,a_t}\tp + \kappa \lambda \I$. We defer the details to \Cref{sec:refined}.
\end{remark}

%% file: sec/analysis.tex
\section{Analysis} \label{sec:analysis}

In this section, we analyze \Cref{thm:regret1}. We first establish a uniform concentration result for the logit estimates and then use it to derive the regret bound. Full proofs are deferred to \Cref{sec:analysis2}.

\subsection{Uniform Concentration of the Logit Estimates}

We first establish a dimension-free high-probability upper bound on $\abs{\v x_{t,a}\tp(\what{\v\theta}_{t-1}^{(s)} - \v\theta^*)}$ simultaneously for all $t\in[T]$, $s\in[S]$, $a\in[K]$.

\para{One-step estimator decomposition}

By the definitions of the initial-point estimator and the one-step corrected estimator in \Cref{eq:theta_bar,eq:theta_hat}, we obtain the following decomposition.

\begin{lemma} \label{lem:decomp}
    For all $\v x\in\R^d$, $t\in[0,T-1]$, $s\in[S]$, we have
    \begin{align*}
        &\abs{\v x\tp(\what{\v\theta}_t^{(s)} - \v\theta^*)} \leq \underbrace{\abs{\v x\tp \br{\br{\m H_t^{(s)}(\bar{\v\theta}_t^{(s)})}^{-1} \sum_{i\in E^{(s)}(t)} \eps_i \v x_{i,a_i}}}}_{\mathrm{(noise)}} + \underbrace{\abs{\v x\tp \br{\m H_t^{(s)}(\bar{\v\theta}_t^{(s)})}^{-1} \lambda \v\theta^*}}_{\mathrm{(bias)}} \\
    &\quad + \underbrace{\abs{\v x\tp \br{\m H_t^{(s)}(\bar{\v\theta}_t^{(s)})}^{-1}\br{\sum_{i\in E^{(s)}(t)} \br{ \dot\mu(\v x_{i,a_i}\tp\check{\v\theta}_{t,i}^{(s)}) - \dot\mu(\v x_{i,a_i}\tp\bar{\v\theta}_t^{(s)})} \v x_{i, a_i} \v x_{i,a_i}\tp }(\v\theta^* - \bar{\v\theta}_t^{(s)})}}_{\mathrm{(remainder)}},
    \end{align*}
    where $\check{\v\theta}_{t,i}^{(s)} = c_i \v\theta^* + (1-c_i) \bar{\v\theta}_t^{(s)}$ is induced by the mean-value theorem for some $c_i\in[0,1]$.
\end{lemma}

\para{Noise term}

A key feature of our one-step estimator is that, in the (noise) term in \Cref{lem:decomp}, the Hessian $\m H_t^{(s)}(\bar{\v\theta}_t^{(s)})$ and the noise $\eps_i$ are built from disjoint sample subsets. This separation allows us to condition on a $\sigma$-field with respect to which the coefficients are measurable, while the noises in $E^{(s)}(t)$ remain conditionally zero-mean and conditionally independent.
Define
\begin{align*}
    \cal H_{s,t} \coloneqq &\sigma \br{\{ \v x_{i,a}\}_{i\in[t],a\in[K]},~\{r_j\}_{j\in \Psi^{(1)}(t)  \cup \cdots \cup \Psi^{(s-1)}(t)}} \\&\lor \sigma \br{\{\v x_{i,a_i}\}_{i\in E^{(s)}(t)}} \lor \sigma \br{\{\v x_{i,a_i}, r_i\}_{i\in P^{(s)}(t)}}.
\end{align*}
Then, for all $s\in[S]$ and $t\in[T]$, conditional on $\cal H_{s,t}$, $\{\eps_i\}_{i\in E^{(s)}(t)}$ are independent and $\{c_i\}_{i\in E^{(s)}(t)}$ are deterministic (see \Cref{lem:cond_ind}). Therefore, unlike the previous approach, 
we can directly apply a scalar-valued concentration inequality (Bernstein inequality, \Cref{lem:bernstein}).

\begin{lemma} \label{lem:noise}
    Fix $\v x \in\R^d, s\in[S], t\in[0,T-1]$. Then with probability at least $1-\delta$, we have
    \begin{align*}
        \mathrm{(noise)} \leq \br{\sqrt{2L_\mu \kappa^2 \log(2/\delta)}  + \frac{\sqrt{\kappa}\log(2/\delta)}{3\sqrt{\lambda}}} \|\v x \|_{(\m V_{E,t}^{(s)})^{-1}}.
    \end{align*}
\end{lemma}

\para{Remainder term}

As in \Cref{eq:diff1,eq:diff2}, controlling the (remainder) term reduces to control
\begin{align*}
    r_t^{(s)} \coloneqq \max_{i\in E^{(s)}(t)} \abs{\v x_{i,a_i}\tp(\bar{\v\theta}^{(s)}_t - \v\theta^*)}.
\end{align*}
Let $\cal E_1$ denote the event that $\|\bar{\v\theta}_t^{(s)}-\v\theta^*\|_{\m V_{P,t}^{(s)}} \leq \beta$ simultaneously for all $s\in[S]$ and $t\in[0,T-1]$. By the well-known self-normalized bound of \citet{abbasi2011improved}, this event holds with high probability (see \Cref{lem:pilot}). The next lemma shows that $r_t^{(s)}$ is uniformly bounded on this event.

\begin{lemma} \label{lem:small_r}
    On the event $\cal E_1$, for all $t\in[0,T-1], s\in[S]$, we have $r_t^{(s)}\leq \beta \sqrt{\tau^{(s)}} \leq 1$.
\end{lemma}
\begin{proof} [Proof of \Cref{lem:small_r}]
The event $\{i\in E^{(s)}(t)\}$ implies that in round $i$, condition (a2) is satisfied, \begin{align} \| \v x_{i, a_i}\|_{(\m V_{P,i-1}^{(s)})^{-1}}^2 \leq \tau^{(s)} \label{eq:smallr1} \end{align} Therefore, on the event $\cal E_1$, we have \begin{align*} r_t^{(s)} &\leq \max_{i\in E^{(s)}(t)} \|\v x_{i,a_i}\|_{(\m V^{(s)}_{P,i-1})^{-1}} \cdot \|\bar{\v\theta}^{(s)}_{t} - \v\theta^*\|_{\m V^{(s)}_{P,i-1}} \tag{Cauchy-Schwarz} \\ &\leq \|\bar{\v\theta}^{(s)}_{t} - \v\theta^*\|_{\m V^{(s)}_{P,t}} \cdot \max_{i\in E^{(s)}(t)} \|\v x_{i,a_i}\|_{(\m V^{(s)}_{P,i-1})^{-1}} \tag{$\m V_{P,t}^{(s)}\succeq \m V_{P,i-1}^{(s)}$, $\forall i\in E^{(s)}(t)\subset[t]$} \\ &\leq \|\bar{\v\theta}^{(s)}_{t} - \v\theta^*\|_{\m V^{(s)}_{P,t}} \cdot \sqrt{\tau^{{(s)}}} \tag{\Cref{eq:smallr1}} \\ &\leq \beta \cdot \sqrt{\tau^{{(s)}}} \tag{$\cal E_1$} \leq 1, \end{align*}
where the last inequality follows from the definition of $\tau^{(s)}$.
\end{proof}

\begin{lemma} \label{lem:remainder_final}
    On the event $\cal E_1$, for all $\v x \in\R^d,t\in[0,T-1],s\in[S]$, $\mathrm{(remainder)}\leq \frac{\alpha}{2} \|\v x\|_{(\m V^{(s)}_{E,t})^{-1}}$.
\end{lemma}

\para{Final bound}

The (bias) term arises from the regularization parameter $\lambda$ and can be bounded straightforwardly, so we omit its proof.
Combining \Cref{lem:decomp,lem:noise,lem:remainder_final}, together with the bound on the (bias) term, yields the following uniform concentration result.

\begin{theorem} \label{thm:concen}
    With probability at least $1-\delta$, simultaneously for all $t\in[T], s\in[S], a\in[K]$, we have $\abs{m_{t,a}^{(s)} - \v x_{t,a}^\top \v\theta^*} \leq w_{t,a}^{(s)}$.
\end{theorem}

\subsection{Regret Analysis} \label{sec:regret_analysis}

We now prove \Cref{thm:regret1}. Let $\cal E$ denote the good event when \Cref{thm:concen} holds. The following lemma is a standard consequence of the Auer framework and are included for completeness.

\begin{lemma} \label{lem:per-round}
    On the good event $\cal E$, the per-round regret is upper bounded by
    \begin{align*}
        \mu(\v x_{t,a_t^*}\tp \v\theta^*) - \mu(\v x_{t,a_t}\tp\v\theta^*) \leq
        \begin{cases} 
    	   8L_\mu 2^{-s_t} & \text{if $a_t$ is selected by condition (a)}, \\ 
            2L_\mu/\sqrt{T} & \text{if $a_t$ is selected by condition (b)}. 
        \end{cases}
    \end{align*}
\end{lemma} 

The next lemma bounds the cardinality of the pilot set.

\begin{lemma} \label{lem:card_P}
    For all $s\in[S]$, we have $\abs{P^{(s)}(T)} \leq 2d\log(1+T/(\kappa\lambda d))(\tau^{(s)})^{-1}$.
\end{lemma}

We are now ready to prove \Cref{thm:regret1}.

\begin{proof} [Proof of \Cref{thm:regret1}]
Let $\Delta_t \coloneqq \mu(\v x_{t,a_t^*}\tp \v\theta^*) - \mu(\v x_{t,a_t}\tp\v\theta^*)$. Then, we have
\begin{align*}
    R_T &=\uterm{\sum_{s=1}^S \sum_{t\in P^{(s)}(T)} \Delta_t}{1} + \uterm{\sum_{s=1}^S \sum_{t\in E^{(s)}(T)} \Delta_t}{2} + \uterm{\sum_{t\in ([T]\setminus (\Psi^{(1)}(T)\cup \dots \cup \Psi^{(S)}(T)))} \Delta_t}{3}.
\end{align*}
For the \term 1
\begin{align*}
    \term 1 &\leq \sum_{s=1}^S \sum_{t\in P^{(s)}(T)} 8 L_\mu 2^{-s_t} \leq 8L_\mu \sum_{s=1}^S \abs{P^{(s)}(T)} 2^{-s} \tag{\Cref{lem:per-round}} \\
    &\leq 8L_\mu \sum_{s=1}^S \frac{2^{-s} 2 d \log (1+T/(\kappa \lambda d))}{\tau^{(s)}} \tag{\Cref{lem:card_P}}\\
    &\leq 8L_\mu \sum_{s=1}^S 2 d \log (1+T/(\kappa \lambda d)) \beta^2 \br{2^{-s} + \sqrt{32L_\mu \kappa d \log(1+T/(\kappa \lambda d))}} \tag{\Cref{eq:tau}}\\
    &\leq 16 L_\mu  d \log (1+T/(\kappa \lambda d)) \beta^2 \br{1 + S\sqrt{32L_\mu \kappa d \log(1+T/(\kappa \lambda d))}}.
\end{align*} 
For the \term 2, we have
\begin{align*}
    \term 2 \leq \sum_{s=1}^S \sum_{t\in E^{(s)}(T)} 8 L_\mu 2^{-s_t} \leq 8L_\mu \sum_{s=1}^S \sum_{t\in E^{(s)}(T)} w_{t,a_t}^{(s)} =8L_\mu \alpha \sum_{s=1}^S \sum_{t\in E^{(s)}(T)} \| \v x_{t,a_t}\|_{(\m V_{E,t-1}^{(s)})^{-1}},
\end{align*}
where the first inequality follows from \Cref{lem:per-round}, and the second inequality follows from the condition (a).
Denote $E^{(s)}(T)$ by $E^{(s)}$, and enumerate its elements in chronological order as $t^{(1)},\dots,t^{(|E^{(s)}|)}$. Let $\m V_0^{(s)} = \kappa \lambda \I$, $\m V^{(s)}_j = \m V^{(s)}_{j-1} + \v x_{t^{(j)}, a_{t^{(j)}}} \v x_{t^{(j)}, a_{t^{(j)}}}\tp$. Then, we can proceed as
\begin{align*}
    \term 2 &\leq 8L_\mu \alpha \sum_{s=1}^S \sum_{j=1}^{|E^{(s)}|} \min \cbr{1, \norm{\v x_{t^{(j)}, a_t^{(j)}}}_{(\m V_{j-1}^{(s)})^{-1}}} \tag{$\lambda\geq1/\kappa$, $\norm{\v x_{t^{(j)},a_{t^{(j)}}}}_{(\m V^{(s)}_{j-1})^{-1}} \leq 1$} \\
    &\leq 8L_\mu \alpha \sqrt{S  \sum_{s=1}^S \abs{E^{(s)}} \sum_{j=1}^{|E^{(s)}|} \min \cbr{1, \norm{\v x_{t^{(j)}, a_t^{(j)}}}_{(\m V_{j-1}^{(s)})^{-1}}^2}} \tag{Cauchy-Schwarz} \\
    &\leq  8L_\mu \alpha  \sqrt{S\sum_{s=1}^S  \abs{E^{(s)}} 2 d \log (1 + T/(\kappa \lambda d))} \tag{\Cref{lem:yadkori}} \\
    &\leq 8L_\mu \alpha \sqrt{2 S T d \log (1 + T/(\kappa \lambda d))}.
\end{align*}
For the \term 3, following from \Cref{lem:per-round}, we have $\term 3 \leq T \times 2L_\mu/\sqrt{T} = 2L_\mu \sqrt{T}$.
Substituting \term 1, \term 2, and \term 3 back completes the proof.
\end{proof}

%% file: sec/experiments.tex
\section{Experiments}

\begin{figure}[t]
    \centering
    \includegraphics[width=\linewidth]{fig/paper_main_figure_4x3_v4.png}
    \caption{Comparison across ambient dimension $d$ and realized geometry. Top row: trajectories of $\log (\det \m V_t / \det \kappa\lambda \I)$ under three regimes, \texttt{high}, \texttt{middle}, and \texttt{low}. Bottom rows: cumulative regret of SupSplitLog, SupCB-GLM, SupLogistic, and DDRTS-GLM on the corresponding instances.}
    \label{fig:1}
\end{figure}

In this section, we empirically evaluate the performance of SupSplitLog. In all experiments, we use the refined version in \Cref{alg:2}, which replaces $\beta$ and $\tau^{(s)}$ with the data-dependent quantities $\beta_t^{(s)}$ and $\tau_t^{(s)}$; see \Cref{sec:refined} for details. We compare SupSplitLog with the three baselines listed in \Cref{table:1}: SupCB-GLM \citep{li2017provably}, SupLogistic \citep{jun2021improved}, and DDRTS-GLM \citep{kim2023double}. For each algorithm, we report the mean cumulative regret with $\pm 1$ standard deviation.

To study the dependence on the ambient feature dimension $d$, we vary both the value of $d$ and the geometry of the incoming features. For each value of $d$, we consider three regimes, labeled \texttt{high}, \texttt{middle}, and \texttt{low}. These regimes correspond to different effective subspace structures of the incoming features and hence different values of $\log \frac{\det \m V_t}{\det \kappa \lambda \I}$, where $\m V_t = \kappa\lambda \I + \sum_{i=1}^t \sum_{a\in[K]} \v x_{i,a} \v x_{i,a}\tp$.

\Cref{fig:1} reports the results. The top row shows the trajectory of $\log \frac{\det \m V_t}{\det \kappa \lambda \I}$ over time under the three regimes, and the bottom rows show the corresponding cumulative regrets. In the \texttt{high} regime, where the incoming features effectively span the ambient space, DDRTS-GLM is competitive. In the \texttt{middle} and \texttt{low} regimes, where the incoming features concentrate on a lower-dimensional structure, SupSplitLog performs most favorably. These results highlight the improved dimension dependence of our method and show that its advantage is most visible when the effective feature dimension is much smaller than the ambient dimension. Full implementation details of the bandit environments and baseline algorithms are deferred to \Cref{sec:exp_details}.

\section{Conclusion and Future Work}

We propose SupSplitLog, which, to the best of our knowledge, is the first algorithm for logistic bandits that achieves $\wtilde{\cal O}(\sqrt{dT})$ regret without any context diversity assumption. The key idea is a sample-splitting design that separates the construction of an initial-point estimator from a one-step correction, yielding a dimension-free high-probability upper bound on the estimation error. This leads to a regret bound with improved dependence on $d$ over previous work.

There are several interesting directions for future work. One is to reduce the $d^{2.5}$ factor in the non-leading term of our regret bound. A sufficiently improved non-leading dependence on $d$ may make it possible to obtain sublinear regret even in regimes where $d$ grows proportionally with $T$. Another direction is to improve the dependence on $\kappa$, which remains a major theme in logistic bandits.

%% file: sec/app_additional_related_work.tex
\section{Additional Related Work} \label{sec:additional_related_work}


Logistic bandits naturally model sequential decision-making problems with binary feedback, and have therefore been studied in a range of application domains. Recent examples include queueing and service systems \citep{kim2024queueing,fu2024efficient,bae2026queue}, as well as preference-based and pairwise-feedback settings \citep{verma2025neural,pasztor2024bandits,bengs2021preference,di2023variance}.

One of the main directions in theoretical research on logistic bandits has focused on infinite-armed generalized linear and logistic bandit problems. 
\citet{filippi2010parametric} initiated this line by establishing a $\tildeO{d\sqrt{T}}$ regret bound for generalized linear bandits. Subsequent work improved the dependence on problem-dependent quantities such as $\kappa$, derived sharper instance-dependent guarantees, and extended the analysis to broader generalized linear settings \citep{faury2020improved,abeille2021instance,faury2022jointly,lee2024unified,sawarni2024generalized,liu2024almost}. There are also several related directions, such as developing jointly efficient online algorithms \citep{jun2017scalable,zhang2023online,zhang2025generalized}, improving the dependence on the norm bound $B$ \citep{lee2024improved,lee2025improved}, designing Thompson sampling-based algorithms \citep{dong2019performance,gouverneur2024information}, extending logistic bandits to multinomial logit models \citep{boudart2026enjoying}, and taking care of unknown reward functions \citep{kang2025single}.

Another direction is to provide improved regret bounds for the finite-armed setting where at each round the agent takes a set of $K$ actions. 
This line of research is also closely related to finite-action linear bandits. In particular, the Auer framework \citep{auer2002using} and its refinements \citep{valko2013finite,chu2011contextual,li2019nearly,li2021tight} provide the basic independent-sample construction on which later finite-armed logistic bandit algorithms are built. For the $K$-armed logistic bandit setting, \cite{li2017provably} came up with an algorithm with a regret bound of order $\wtilde{O}(\sqrt{dT})$. Later, \cite{jun2021improved} refined the algorithm and improved the dependence on $\kappa$ on the regret bound. More recently, \cite{kim2023double} developed a Thompson sampling-based algorithm, which improved practicality and numerical performance. However, as mentioned earlier, discussed, these works heavily rely on context-diversity assumptions. We note that context-diversity assumptions also appear in several related generalized linear settings \citep{wu2020stochastic,papini2021leveraging,goyal2025efficient}.

In fact, context-diversity assumptions are the main hurdle in extending the $K$-armed logistic bandit frameworks to high-dimensional settings. For $K$-armed linear bandits, we may obtain a $\wtilde{O}(\sqrt{dT})$ regret bound without a context-diversity assumption, based on which we can extend such result to richer high-dimensional models such as kernerlized and neural bandits \citep{zhou2020neural,kassraie2022neural}. In contrast, to our knowledge, existing kernelized and neural extensions of logistic bandit models do not admit an analogous $\widetilde O(\sqrt{dT})$ regret bound. The obstacle is that the context-diversity assumptions used in the $K$-armed setting do not extend directly to kernelized and neural settings without incurring direct dependence on the feature dimension $d$ \citep{verma2025neural,bae2025neural,akhavan2025bernstein,metelli2025generalized}.


Lastly, we note that there are other finite-action settings with similar feedback models, including dueling bandits \citep{saha2021optimal}, multinomial logit bandits \citep{oh2021multinomial}, and the fixed-arm set setting \citep{mason2022experimental}. However, these works also assume some richness and diversity conditions or consider more restricted settings. Taken together, these directions show that removing context diversity assumptions while retaining $\wtilde{\cal O}(\sqrt{dT})$ regret in finite-armed logistic bandits has remained open prior to our work.

%% file: sec/app_algorithm2.tex
\section{Refined Algorithm for Data-Dependent Regret} \label{sec:refined}

\begin{algorithm}[htbp]
\caption{SupSplitLog with data-dependent regret}
\label{alg:2}
\begin{algorithmic}[1]
\Input $d,T,K,B,\kappa,\lambda,\delta$
\Init $S\gets\lfloor\log_2T\rfloor$, $\alpha$ as in \Cref{eq:alpha}
\Init For $s\in[S]$,~~~~$\Psi^{(s)}\gets\emptyset$, $P^{(s)}\gets\emptyset$, $E^{(s)}\gets\emptyset$, $\bar{\v\theta}^{(s)}_0\gets \v 0$
\For{$t=1,\dots,T$}
  \State Initialize $A^{(1)}\gets[K]$, $s\gets 1$, $a_t\gets \mathtt{None}$
  \While{$a_t=\mathtt{None}$}
    \State $\m V^{(s)}_{P,t-1} \gets \kappa \lambda \I + \sum_{i\in P^{(s)}(t-1)} \v x_{i,a_i} \v x_{i, a_i}\tp$
    \State $\m V^{(s)}_{E,t-1} \gets \kappa \lambda \I + \sum_{i\in E^{(s)}(t-1)} \v x_{i,a_i} \v x_{i, a_i}\tp$
    \State Update $\bar{\v\theta}_{t-1}^{(s)}$ as in \Cref{eq:theta_bar},~~~~$\what{\v\theta}_{t-1}^{(s)}$ as in \Cref{eq:theta_hat}
    \State \note{Update $\beta_{t-1}^{(s)}$ as in \Cref{eq:beta2},~~~~$\tau_{t-1}^{(s)}$ as in \Cref{eq:tau2}}
    \State For $a\in A^{(s)}$,~~~~compute $m^{(s)}_{t,a}\gets \v x_{t,a}\tp \what{\v\theta}_{t-1}^{(s)}$,~~~~$w^{(s)}_{t,a}\gets \alpha \|\v x_{t,a}\|_{(\m V_{E,t-1}^{(s)})^{-1}}$
    \If{$\exists a\in A^{(s)}\st w^{(s)}_{t,a}>2^{-s}$} \Comment{condition (a)}
      \State Choose $a_t\gets \argmax_{a\in A^{(s)}} w^{(s)}_{t,a}$,~~~~receive $r_t$
      \State $\Psi^{(s)}\leftarrow \Psi^{(s)} \cup \{t\}$
      \If {\note{$\|\v x_{t,a_t}\|_{(\m V_{P,t-1}^{(s)})^{-1}}^2 > \tau_{t-1}^{(s)}$}} \Comment{condition (a1)}
        \State $P^{(s)}\leftarrow P^{(s)} \cup \{t\}$
      \Else \Comment{condition (a2)}
        \State $E^{(s)}\leftarrow E^{(s)} \cup \{t\}$
      \EndIf
    \ElsIf{$\forall a\in A^{(s)}~,~ w^{(s)}_{t,a}\le 1/\sqrt{T}$} \Comment{condition (b)}
      \State Choose, $a_t\gets\argmax_{a\in A^{(s)}} m^{(s)}_{t,a}$,~~~~receive $r_t$
    \Else \Comment{condition (c)}
      \State $A^{(s+1)}\gets\{a\in A^{(s)} \st m^{(s)}_{t,a}\geq \max_{a\in A^{(s)}}m^{(s)}_{t,a}-2\cdot 2^{-s}\}$
      \State $s\gets s+1$
    \EndIf
  \EndWhile
\EndFor
\end{algorithmic}
\end{algorithm}

As mentioned in \Cref{remark:data-dependent}, in this section we present a refined version of \Cref{alg:1}, illustrated in \Cref{alg:2}, which yields a data-dependent regret bound.

\begin{align}
     \tau_t^{(s)} &= \frac{1}{(\beta_t^{(s)})^2} \min\cbr{1,~ 2^{-s} \br{32 L_{\mu} \kappa \log\br{ \frac{\det \m V_{E,t}^{(s)}}{\det \kappa \lambda \I}}}^{-1/2}} \label{eq:tau2}  \\
     \beta_t^{(s)} &= \frac{\kappa}{\sqrt{2}} \br{\log \br{ \frac{\det \m V_{P,t}^{(s)}}{\det \kappa \lambda \I}} +  \log (2S/\delta)}^{1/2} + B\sqrt{\kappa \lambda}, \label{eq:beta2}
\end{align}

In addition, let $\m V_t = \kappa \lambda \I + \sum_{i=1}^t \v x_{i,a_i} \v x_{i,a_i}\tp$ be the design matrix formed by all features selected up to round $t$, and define $\beta_t = \frac{\kappa}{\sqrt{2}} \br{\log \br{ \frac{\det \m V_{t}}{\det \kappa \lambda \I}} +  \log (2S/\delta)}^{-1/2} + B\sqrt{\kappa \lambda}$.
Then, $\m V_t \succeq \m V_{P,t}^{(s)}$, $\m V_t \succeq \m V_{E,t}^{(s)}$, and $\beta_t\geq \beta_t^{(s)}$ for all $s\in[S]$ and $t\in[T]$.

\begin{theorem} \label{thm:regret2}
    Define a data-dependent complexity measure $\wtilde d$ as
    \begin{align}
        \wtilde d \coloneqq \log\br{\frac{\det \m V_T}{\det \kappa \lambda \I}} = \log \det \br{\I + \frac{1}{\kappa\lambda}\sum_{t=1}^T \v x_{t,a_t} \v x_{t,a_t}\tp}. \label{eq:logdet}
    \end{align}
    Then, with probability at least $1-\delta$, the regret of \Cref{alg:2} is upper bounded by 
    \begin{align*}
        R_T \leq  16 L_\mu  \wtilde d \beta_T^2 \br{1 + S\sqrt{32L_\mu \kappa \wtilde d}} + 8L_\mu \alpha \sqrt{2 S T \wtilde d} + 2L_\mu \sqrt{T}.
    \end{align*}
\end{theorem}

\begin{remark}[Remark on $\wtilde d$]
This log-determinant quantity $\wtilde d$ is a data-dependent measure of the realized feature geometry. In particular,
under $\|\v x_{t,a_t}\|_2 \le 1$, we have $\wtilde d \leq d\log(1+T/(\kappa\lambda d))$, so $\wtilde d$ is always upper bounded by the ambient-dimension scale up to logarithmic factors, while it can be
substantially smaller when the selected features are concentrated on a low-dimensional subspace and have small norm.

This log-determinant quantity is also standard in kernelized and Gaussian-process bandits \citep{srinivas2012information,chowdhury2017kernelized,vakili2021information}: an analogous information-gain quantity is given  by $\frac{1}{2}\log\det(I+\lambda^{-1}K_T)$, where $K_T \in \mathbb{R}^{T\times T}$ is the kernel Gram matrix of the selected points, defined by $(K_T)_{ij}=k(\v x_{i,a_i},\v x_{j,a_j})$ for $i,j\in[T]$. Such a quantity is essential in kernelized settings because the feature dimension induced by the kernel can be very large or even infinite, so regret bounds stated directly in terms of $d$ become unavailable or uninformative.
\end{remark}

\subsection{Proof of \Cref{thm:regret2}}

We first collect the lemmas that can be obtained from the previous analysis with only minor modifications. Let $\cal E_1'$ denote the event on which \Cref{lem:pilot2} holds.

\begin{lemma} \label{lem:pilot2}
    With probability at least $1-\delta/2$, simultaneously for all $s\in[S], t\in[0,T-1]$, we have $\norm{\bar{\v\theta}^{(s)}_t - \v\theta^*}_{\m V_{P,t}^{(s)}} \leq \beta_t$.
\end{lemma}

\begin{lemma} \label{lem:small_r2}
    On the event $\cal E_1'$, for all $t\in[0,T-1], s\in[S]$, we have $r_t^{(s)}\leq \beta_t^{(s)} \sqrt{\tau_t^{(s)}} \leq 1$.
\end{lemma}

\begin{lemma} \label{lem:e_card2}
    For all $t\in [0,T-1], s\in [S]$, we have $|E^{(s)}(t)| \leq 2 \alpha^2 4^s \log \br{\frac{\det \m V_{E,t}^{(s)}}{\det \kappa \lambda \I}}$.
\end{lemma}

We omit the proofs of \Cref{lem:pilot2,lem:small_r2,lem:e_card2}, since they follow directly from the proofs of \Cref{lem:pilot,lem:small_r,lem:e_card}, respectively.

\begin{lemma} \label{lem:remainder_final2}
    On the event $\cal E_1'$, for all $\v x \in\R^d,t\in[0,T-1],s\in[S]$, we have $\mathrm{(remainder)}\leq \frac{\alpha}{2} \|\v x\|_{(\m V^{(s)}_{E,t})^{-1}}$.
\end{lemma}

\begin{proof} [Proof of \Cref{lem:remainder_final2}]
We fix $t$ and $s$ and drop the relevant notations for simplicity.
From the result of \Cref{lem:remainder},
\begin{align*}
        &\abs{\v x\tp \m H^{-1}(\bar{\v\theta})\br{\sum_{i\in E} \br{ \dot\mu(\v x_{i,a_i}\tp\check{\v\theta}_i) - \dot\mu(\v x_{i,a_i}\tp\bar{\v\theta})} \v x_{i, a_i} \v x_{i,a_i}\tp }(\v\theta^* - \bar{\v\theta})} \\
        &\quad \leq r(e^{r} - 1) \sqrt{|E|} \sqrt{L_\mu \kappa} \|\v x\|_{\m V_{E}^{-1}} \\
        &\quad \leq r(e^r - 1) \alpha 2^s \sqrt{2  \log \br{\frac{\det \m V_{E,t}^{(s)}}{\det \kappa \lambda \I}}} \sqrt{L_{\mu}\kappa} \|\v x\|_{\m V_{E}^{-1}} \tag{\Cref{lem:e_card2}} \\
        &\quad \leq r^2 \alpha 2^{s+1} \sqrt{2  \log \br{\frac{\det \m V_{E,t}^{(s)}}{\det \kappa \lambda \I}}} \sqrt{L_{\mu}\kappa} \|\v x\|_{\m V_{E}^{-1}} \tag{\Cref{lem:small_r2}, $e^r-1\leq 2r$, $\forall r\leq1$} \\
        &\quad \leq \alpha \tau_t^{(s)} (\beta_t^{(s)})^2 2^{s+1} \sqrt{2  \log \br{\frac{\det \m V_{E,t}^{(s)}}{\det \kappa \lambda \I}}} \sqrt{L_{\mu}\kappa} \|\v x\|_{\m V_{E}^{-1}} \tag{\Cref{lem:small_r2}} \\
        &\quad \leq \frac{\alpha}{2} \|\v x\|_{\m V_{E}^{-1}}, 
    \end{align*}
    where the last inequality follows from the definition of $\tau_t^{(s)}$.
\end{proof}

Lastly, we state the analogue of \Cref{lem:card_P} for the pilot set. Since its proof is identical to that of \Cref{lem:card_P}, we omit it.

\begin{lemma} \label{lem:card_P2}
    For all $s\in[S]$, we have $\abs{P^{(s)}(T)} \leq 2\log \br{\frac{\det \m V_{P, T}^{(s)}}{\det \kappa \lambda \I}} (\tau_T^{(s)})^{-1}$.
\end{lemma}

With these lemmas in hand, the proof of \Cref{thm:regret2} proceeds exactly as that of \Cref{thm:regret1}, with the ambient-dimension terms replaced by the corresponding log-determinant quantities.

\begin{proof} [Proof of \Cref{thm:regret2}]

We have, 
\begin{align*}
    R_T &=\uterm{\sum_{s=1}^S \sum_{t\in P^{(s)}(T)} \sqbr{\mu(\v x_{t,a_t^*}\tp \v\theta^*) - \mu(\v x_{t,a_t}\tp\v\theta^*)}}{1} + \uterm{\sum_{s=1}^S \sum_{t\in E^{(s)}(T)} \sqbr{\mu(\v x_{t,a_t^*}\tp \v\theta^*) - \mu(\v x_{t,a_t}\tp\v\theta^*)}}{2} \\
    &\quad + \uterm{\sum_{t\in ([T]\setminus (\Psi^{(1)}(T)\cup \dots \cup \Psi^{(S)}(T)))} \sqbr{\mu(\v x_{t,a_t^*}\tp \v\theta^*) - \mu(\v x_{t,a_t}\tp\v\theta^*)}}{3}.
\end{align*}
For the \term 1
\begin{align*}
    \term 1 &\leq \sum_{s=1}^S \sum_{t\in P^{(s)}(T)} 8 L_\mu 2^{-s_t} \leq 8L_\mu \sum_{s=1}^S \abs{P^{(s)}(T)} 2^{-s} \tag{\Cref{lem:per-round}} \\
    &\leq 8L_\mu \sum_{s=1}^S 2^{-s} 2\log \br{\frac{\det \m V_{P, T}^{(s)}}{\det \kappa \lambda \I}} (\tau_T^{(s)})^{-1} \tag{\Cref{lem:card_P2}}\\
    &\leq 8L_\mu \sum_{s=1}^S 2\log \br{\frac{\det \m V_{P, T}^{(s)}}{\det \kappa \lambda \I}} (\beta_T^{(s)})^2 \br{2^{-s} + \br{32 L_{\mu} \kappa \log\br{ \frac{\det \m V_{E,t}^{(s)}}{\det \kappa \lambda \I}}}^{1/2}} \tag{\Cref{eq:tau2}}\\
    &\leq 16 L_\mu  \wtilde d \beta_T^2 \br{1 + S\sqrt{32L_\mu \kappa \wtilde d}}.
\end{align*} 
For the \term 2,
\begin{align*}
    \term 2 &\leq \sum_{s=1}^S \sum_{t\in E^{(s)}(T)} 8 L_\mu 2^{-s_t} \tag{\Cref{lem:per-round}} \\
    &\leq 8L_\mu \sum_{s=1}^S \sum_{t\in E^{(s)}(T)} w_{t,a_t}^{(s)} =8L_\mu \alpha \sum_{s=1}^S \sum_{t\in E^{(s)}(T)} \| \v x_{t,a_t}\|_{(\m V_{E,t-1}^{(s)})^{-1}} \tag{condition (a)}.
\end{align*}
Denote $E^{(s)}(T)$ by $E^{(s)}$, and enumerate its elements in chronological order as $t^{(1)},\dots,t^{(|E^{(s)}|)}$. Let $\m V_0^{(s)} = \kappa \lambda \I$, $\m V^{(s)}_j = \m V^{(s)}_{j-1} + \v x_{t^{(j)}, a_{t^{(j)}}} \v x_{t^{(j)}, a_{t^{(j)}}}\tp$. Then, we can proceed as
\begin{align*}
    \term 2 &\leq 8L_\mu \alpha \sum_{s=1}^S \sum_{j=1}^{|E^{(s)}|} \min \cbr{1, \norm{\v x_{t^{(j)}, a_t^{(j)}}}_{(\m V_{j-1}^{(s)})^{-1}}} \tag{$\lambda\geq1/\kappa$, $\norm{\v x_{t^{(j)},a_{t^{(j)}}}}_{(\m V^{(s)}_{j-1})^{-1}} \leq 1$} \\
    &\leq 8L_\mu \alpha \sqrt{S  \sum_{s=1}^S \abs{E^{(s)}} \sum_{j=1}^{|E^{(s)}|} \min \cbr{1, \norm{\v x_{t^{(j)}, a_t^{(j)}}}_{(\m V_{j-1}^{(s)})^{-1}}^2}} \tag{Cauchy-Schwarz} \\
    &\leq  8L_\mu \alpha  \sqrt{S\sum_{s=1}^S  \abs{E^{(s)}} 2 \log\br{ \frac{\det \m V_{E,T}^{(s)}}{\det \kappa \lambda \I}}} \tag{\Cref{lem:yadkori}} \\
    &\leq 8L_\mu \alpha \sqrt{2 S T \wtilde d}.
\end{align*}
For the \term 3,
\begin{align*}
    \term 3 \leq T \times 2L_\mu/\sqrt{T} = 2L_\mu \sqrt{T} \tag{\Cref{lem:per-round}}.
\end{align*}
Substituting \term 1, \term 2, and \term 3 back completes the proof.
\end{proof}

%% file: sec/app_exp.tex
\section{Experimental details} \label{sec:exp_details}

This section provides implementation details for the synthetic experiment in \Cref{fig:1}. All experiments were conducted on a server equipped with an AMD EPYC 9354 32-Core Processor, 251 GiB of RAM, and one NVIDIA RTX A6000 GPU.

\para{Environment}
We use a synthetic logistic bandit environment in which contexts are generated at initialization. The generator explicitly enforces $\|\v x_{t,a}\|_2 \leq 1$, $\|\v\theta^*\|_2 \leq B$, and $\dot\mu(\v x_{t,a}\tp\v\theta^*) \geq 1/\kappa$.
For \Cref{fig:1}, we fix $T=2000$, $K=5$, $\kappa=20$, $\lambda=1$, $B=1$, and vary the ambient dimension over $d\in\{3,20,100\}$. For each value of $d$, we consider three geometry regimes, labeled \texttt{high}, \texttt{middle}, and \texttt{low}. These regimes are implemented by changing the effective subspace dimension $r$ and the norm range of the incoming features:
\begin{align*}
\texttt{low}:&\quad r=2,~\|\v x\|_2\in[0,0.05],\\
\texttt{middle}:&\quad r=10,~\|\v x\|_2\in[0.3,0.5],\\
\texttt{high}:&\quad r=d,~\|\v x\|_2\in[0.8,1.0].
\end{align*}
Thus, the three regimes correspond to different effective subspace structures and lead to different values of the log-determinant quantity shown in \Cref{fig:1}. All experiments are repeated over $10$ independent runs, and the plots report the mean performance with $\pm 1\sigma$ error bars.

\para{Baseline algorithms}
We compare the refined version of our method in \Cref{alg:2} with three baselines: SupCB-GLM \citep{li2017provably}, SupLogistic \citep{jun2021improved}, and DDRTS-GLM \citep{kim2023double}. During implementation, \Cref{alg:2} maintains pilot and estimation buckets at each level, computes an initial-point estimator on the pilot data, and then applies a one-step Newton correction on the estimation data. The thresholds $\tau_t^{(s)}$ are updated adaptively using the current log-determinant statistics.

SupCB-GLM and SupLogistic are both implemented with independent level-wise buckets. In both cases, each level has its own warm-up size $\tau=\sqrt{dT}$, and samples are not reused across levels. DDRTS-GLM is implemented as a practical Thompson-sampling-style baseline that repeatedly fits a regularized logistic estimator, forms a Hessian-based covariance matrix, and samples actions using Monte Carlo argmax probabilities.

All methods use the same projected gradient descent routine for fitting regularized logistic objectives, and we use $20$ gradient steps with learning rate $0.3$. For DDRTS-GLM, we use one Monte Carlo sample to approximate argmax probabilities.

%% file: sec/app_proofs.tex
\section{Deferred Proofs for \Cref{sec:analysis}} \label{sec:analysis2}

\subsection{Proof of \Cref{lem:decomp}} \label{sec:lem:decomp}
\begin{proof}
Consider the term $\what{\v\theta}_{t}^{(s)} - \v\theta^*$. In this section, we fix $t$ and $s$ and drop the notations for simplicity. Then, we have
\begin{align*}
    \what{\v\theta} - \v\theta^* & = \bar{\v\theta} + \m H^{-1}(\bar{\v\theta}) \v g(\bar{\v\theta}) - \v\theta^* \\
    &= \bar{\v\theta} + \m H^{-1}(\bar{\v\theta}) \br{\sum_{i\in E} \br{r_i - \mu(\v x_{i,a_i}\tp \bar{\v\theta})}\v x_{i,a_i} - \lambda \bar{\v\theta}} -\v\theta^* 
\end{align*}
By the definition,
\begin{align*}
    r_i - \mu(\v x_{i,a_i}\tp \bar{\v\theta})  &= \eps_i + \mu(\v x_{i,a_i}\tp \v\theta^*) - \mu(\v x_{i,a_i}\tp \bar{\v\theta}) \\
    &=\eps_i + \dot\mu(\v x_{i,a_i}\tp \check{\v\theta}_i) \v x_{i,a_i}\tp (\v\theta^* - \bar{\v\theta}),
\end{align*}
where the last equality holds by the mean-value theorem with some $c_i>0$ and $\check{\v\theta}_i = c_i\v\theta^* + (1-c_i)\bar{\v\theta}$. Substituting this result back gives
\begin{align*}
     \what{\v\theta} - \v\theta^* &= \m H^{-1}(\bar{\v\theta}) \sum_{i\in E} \eps_i \v x_{i,a_i} + \m H^{-1}(\bar{\v\theta})\br{\sum_{i\in E} \dot\mu(\v x_{i,a_i}\tp\check{\v\theta}_i) \v x_{i, a_i} \v x_{i,a_i}\tp (\v\theta^* - \bar{\v\theta})} - \m H^{-1}(\bar{\v\theta}) \lambda \bar{\v\theta} - (\v\theta^* -\bar{\v\theta}) \\
     &= \m H^{-1}(\bar{\v\theta}) \sum_{i\in E} \eps_i \v x_{i,a_i} + \m H^{-1}(\bar{\v\theta})\br{\sum_{i\in E} \br{ \dot\mu(\v x_{i,a_i}\tp\check{\v\theta}_i) - \dot\mu(\v x_{i,a_i}\tp\bar{\v\theta})} \v x_{i, a_i} \v x_{i,a_i}\tp (\v\theta^* - \bar{\v\theta})}    \\
     &\quad + \uterm{\m H^{-1}(\bar{\v\theta})\br{\sum_{i\in E} \dot\mu(\v x_{i,a_i}\tp\bar{\v\theta}) \v x_{i, a_i} \v x_{i,a_i}\tp (\v\theta^* - \bar{\v\theta})} - \m H^{-1}(\bar{\v\theta}) \lambda \bar{\v\theta} - (\v\theta^* -\bar{\v\theta})}{1}
\end{align*}
For the \term 1,
\begin{align*}
    \term 1 &= \m H^{-1}(\bar{\v\theta}) (\m H(\bar{\v\theta}) - \lambda\I) (\v\theta^* - \bar{\v\theta}) - \m H^{-1}(\bar{\v\theta}) \lambda \bar{\v\theta} - (\v\theta^* - \bar{\v\theta}) \\
    &= - \m H^{-1}(\bar{\v\theta}) \lambda (\v\theta^* -\bar{\v\theta}) - \m H^{-1}(\bar{\v\theta}) \lambda \bar{\v\theta} \\
    & = - \m H^{-1}(\bar{\v\theta}) \lambda \v\theta^*.
\end{align*}
Now, plugging this result back, and compute $|\v x\tp (\what{\v\theta} - \v\theta^*)|$ and split the term with triangle inequality completes the proof.    
\end{proof}

\subsection{Proof of \Cref{lem:noise}} \label{sec:lem:noise}

In this section, we fix $t$ and $s$ and drop the notations for simplicity.
Denote
\begin{align*}
    p_i = \v x\tp \m H^{-1}(\bar{\v\theta}) \v x_{i,a_i}, \quad \forall i\in E.
\end{align*}
Then, we have
\begin{align}
    \abs{\v x\tp \br{\m H^{-1}(\bar{\v\theta}) \sum_{i\in E} \eps_i \v x_{i,a_i}}} = \abs{ \sum_{i\in E} p_i \eps_i}. \label{eq:noise0}
\end{align}
We want to apply the Bernstein inequality to \Cref{eq:noise0}. To apply this, we need to show that being conditioned on a proper filtration, (i) $\{\eps_i\}_{i\in E}$ are independent, and (ii) $\{p_i\}_{i\in E}$ are fixed.

First, define a filtration
\begin{align*}
    \cal G_{s,t} \coloneqq \sigma \br{\{ \v x_{i,a}\}_{i\in[t],a\in[K]},~\{r_j\}_{j\in \Psi^{(1)}(t) \cup \Psi^{(2)}(t) \cup \cdots \cup \Psi^{(s-1)}(t)}}.
\end{align*}

\begin{lemma} [Level-$s$ fixed design, \citep{auer2002using}] \label{lem:fixed}
    For all $s\in[S],t\in[T]$, $\Psi^{(s)}(t)$ and $\{x_{i,a_i}\}_{i\in\Psi^{(s)}(t)}$ are $\cal G_{s,t}$-measurable. Therefore, conditional on $\cal G_{s,t} \lor \sigma(\{\v x_{i,a_i}\}_{i\in\Psi^{(s)}(t)})$, $\{r_i\}_{i\in\Psi^{(s)}(t)}$ are independent.
\end{lemma}

\begin{proof}
    Consider any $i\in[t]$. The event $\{i\in\Psi^{(s)}(t)\}$ implies that in round $i$, the algorithm (i) reaches level $s$ and then (ii) falls in to condition (a). These two requirements, (i) and (ii), only depend on arm sets $\leq t$ i.e., $\{ \v x_{i,a}\}_{i\in[t],a\in[K]}$, and the rewards of levels $<s$ i.e., $r_j\in\Psi^{(1)}(t) \cup \Psi^{(2)}(t) \cup \cdots \cup \Psi^{(s-1)}(t)$. This means that $\Psi^{(s)}(t)$ and $\{x_{i,a_i}\}_{i\in\Psi^{(s)}(t)}$ are $\cal G_{s,t}$-measurable.
    Therefore, conditional on $\cal G_{s,t} \lor \sigma(\{\v x_{i,a_i}\}_{i\in\Psi^{(s)}(t)})$, $\{r_i\}_{i\in\Psi^{(s)}(t)}$ are independent Bernoulli random variables. 
\end{proof}

\begin{proposition} \label{prop:fixed}
    For all $s\in[S], t\in[T]$, $P^{(s)}(t)$ and $E^{(s)}(t)$ are also $\cal G_{s,t}$-measurable.
\end{proposition}

\begin{proof}
    Consider any $i\in [t]$ and the event $\{i\in P^{(s)}(t)\}$. The event implies that $\{i\in\Psi^{(s)}(t)\}$, which is $\cal G_{s,t}$-measurable by \Cref{lem:fixed}, and that the condition (a1) holds, which is $\|\v x_{i,a_i}\|_{(\m V^{(s)}_{P,i-1})^{-1}}^2 > \tau^{(s)}$. Since whether such condition holds only depends on $\{\v x_{i,a_i}\}_{i\in\Psi^{(s)}(t)}$, it is also $\cal G_{s,t}$-measurable. Same applies to the event $\{i\in E^{(s)}(t)\}$. 
\end{proof}

Now, we establish (i) and (ii) conditioning on a new filtration 
\begin{align*}
    \cal H_{s,t} \coloneqq \cal G_{s,t} \lor \sigma \br{\{\v x_{i,a_i}\}_{i\in E}} \lor \sigma \br{\{x_{i,a_i}, r_i\}_{i\in P}}.
\end{align*}

\begin{lemma} \label{lem:cond_ind}
    For all $s\in[S],t\in[T]$, conditional on $\cal H_{s,t}$, $\{\eps_i\}_{i\in E^{(s)}(t)}$ are independent, and $\{p_i\}_{i\in E^{(s)}(t)}$ are deterministic.
\end{lemma}

\begin{proof}
For (i), by \Cref{lem:fixed,prop:fixed}, $\{r_i\}_{i\in P\cup E}$ are independent, conditioned on $\cal G_{s,t}\lor \sigma \br{\{\v x_{i,a_i}\}_{i\in E}}\lor \sigma \br{\{\v x_{i,a_i}\}_{i\in P}}$. Therefore, further conditioning on $\sigma(\{r_i\}_{i\in P})$, which is same as $\cal H_{s,t}$, does not affect the independence on $\{r_i\}_{i\in E}$. Also the definition of $\eps_i=r_i - \mu(\v x_{i,a_i}\tp\v\theta^*)$ implies that $\{\eps_i\}_{i\in E}$ are independent.
Next, for (ii), $p_i$ consists of $\{\v x_{i,a_i}\}_{i\in E}$, which is $\cal H_{s,t}$-measurable, and $\m H(\bar{\v\theta})$. From the definition of $\m H(\bar{\v\theta})$, it consists of $\{\v x_{i,a_i}\}_{i\in E}$ and $\bar{\v\theta}$ where $\bar{\v\theta}$ is determined by $\{\v x_{i,a_i}, r_i\}_{i\in P}$. Therefore, $p_i$ is $\cal H_{s,t}$-measurable.
\end{proof}

Now we are ready to start the proof.

\begin{proof} [Proof of \Cref{lem:noise}]
We prepare to apply the Bernstein inequality to \Cref{eq:noise0}.
We start by upper bounding $\max_{i\in E} |p_i \eps_i|$ and $\sum_{i\in E}\Var(p_i \eps_i\mid\cal H_{s,t})$: For $\max_{i\in E} |p_i \eps_i|$, we have
\begin{align}
    \max_{i\in E} |p_i \eps_i| &\leq \max_{i \in E} |\v x\tp \m H^{-1}(\bar{\v\theta}) \v x_{i,a_i}| \nonumber \\
    &\leq \max_{i\in E} \|\v x\tp \m H^{-1/2}(\bar{\v\theta})\|_2 \cdot \|\m H^{-1/2}(\bar{\v\theta}) \|_2 \cdot \|\v x_{i,a_i}\|_2 \tag{Cauchy-Schwarz} \\
    &\leq \frac{1}{\sqrt{\lambda}} \|\v x\|_{\m H^{-1}(\bar{\v\theta})} \tag{$\|\v x_{i,a_i}\|_2\leq 1$, $\lambda_{\min}(\m H(\bar{\v\theta}))\geq\lambda$} \\
    &\leq \sqrt{\frac{\kappa}{\lambda}} \|\v x \|_{\m V^{-1}_{E}} \label{eq:noise1},
\end{align}
where the last inequality holds since $\bar{\v\theta}\in\Theta$, and $\m V_E \preceq \kappa \m H(\bar{\v\theta})$.

For $\sum_{i\in E}\Var(p_i \eps_i \mid \cal H_{s,t})$, we have
\begin{align}
    \sum_{i\in E}\Var(p_i \eps_i \mid \cal H_{s,t}) & = \sum_{i\in E} \E[(p_i \eps_i)^2 \mid \cal H_{s,t}]  \tag{$\E[\eps_i\mid \cal H_{s,t}]=0$}\\
    &= \sum_{i\in E} \v x\tp \m H^{-1}(\bar{\v\theta}) \v x_{i,a_i} \v x_{i,a_i}\tp \m H^{-1}(\bar{\v\theta}) \v x \cdot \E[\eps_i^2 \mid \cal H_{s,t}] \nonumber \\
    &\leq L_\mu  \sum_{i\in E} \v x\tp \m H^{-1}(\bar{\v\theta}) \v x_{i,a_i} \v x_{i,a_i}\tp \m H^{-1}(\bar{\v\theta}) \v x \tag{\Cref{ass:kappa}} \\
    &= L_\mu   \v x\tp \m H^{-1}(\bar{\v\theta}) \br{\sum_{i\in E} \v x_{i,a_i} \v x_{i,a_i}\tp} \m H^{-1}(\bar{\v\theta}) \v x \nonumber \\
    &\leq L_\mu   \v x\tp \m H^{-1}(\bar{\v\theta}) \kappa \m H(\bar{\v\theta}) \m H^{-1}(\bar{\v\theta}) \v x \tag{$\m V_E \preceq \kappa \m H(\bar{\v\theta})$}  \\
    &\leq L_\mu \kappa^2 \|\v x\|_{\m V_E^{-1}}^2. \label{eq:noise2}
\end{align}
Applying Bernstein inequality (\Cref{lem:bernstein}) with
\begin{align*}
    b\leftarrow \sqrt{\frac{\kappa}{\lambda}} \|\v x \|_{\m V^{-1}_{E}},\quad V \leftarrow L_\mu \kappa^2 \|\v x\|_{\m V_E^{-1}}^2, \quad L\leftarrow \log(2/\delta) \tag{\Cref{eq:noise1,eq:noise2}},
\end{align*}
with probability at least $1-\delta$,
\begin{align*}
    \abs{\sum_{i\in E} p_i \eps_i} \leq \sqrt{2L_\mu \kappa^2 \log(2/\delta)} \|\v x\|_{\m V^{-1}_E} + \frac{\sqrt{\kappa}\log(2/\delta)}{3\sqrt{\lambda}} \|\v x\|_{\m V^{-1}_E},
\end{align*}
which completes the proof.
\end{proof}

\subsection{Proof of \Cref{lem:remainder_final}} \label{sec:lem:remainder_final}

Before starting the proof, we introduce two technical lemmas:
\begin{lemma} \label{lem:remainder}
    For all $\v x \in\R^d$, $t\in[0,T-1]$, $s\in[S]$, 
    \begin{align*}
        &\abs{\v x\tp \br{\m H_t^{(s)}(\bar{\v\theta}_t^{(s)})}^{-1}\br{\sum_{i\in E^{(s)}(t)} \br{ \dot\mu(\v x_{i,a_i}\tp\check{\v\theta}_{t,i}^{(s)}) - \dot\mu(\v x_{i,a_i}\tp\bar{\v\theta}_t^{(s)})} \v x_{i, a_i} \v x_{i,a_i}\tp }(\v\theta^* - \bar{\v\theta}_t^{(s)})} \\
        &\quad \leq r_t^{(s)}(e^{r_t^{(s)}} - 1) \sqrt{|E^{(s)}(t)|} \sqrt{L_\mu \kappa} \|\v x\|_{(\m V^{(s)}_{E,t})^{-1}}.
    \end{align*}
\end{lemma}

\begin{proof}
    The proof is deferred to \Cref{sec:lem:remainder}.
\end{proof}

\begin{lemma} \label{lem:e_card}
    For all $t\in [0,T-1], s\in [S]$, we have $|E^{(s)}(t)| \leq 2 \alpha^2 d 4^s \log(1+T/(\kappa\lambda d))$.
\end{lemma}

\begin{proof}
    The proof is deferred to \Cref{sec:lem:e_card}.
\end{proof}

Now we are ready to start the proof.
In this section, we fix $t$ and $s$ and drop the notations  for simplicity.

\begin{proof} [Proof of \Cref{lem:remainder_final}]
We can start from the result of \Cref{lem:remainder}, which is
\begin{align*}
        &\abs{\v x\tp \m H^{-1}(\bar{\v\theta})\br{\sum_{i\in E} \br{ \dot\mu(\v x_{i,a_i}\tp\check{\v\theta}_i) - \dot\mu(\v x_{i,a_i}\tp\bar{\v\theta})} \v x_{i, a_i} \v x_{i,a_i}\tp }(\v\theta^* - \bar{\v\theta})} \\
        &\quad \leq r(e^{r} - 1) \sqrt{|E|} \sqrt{L_\mu \kappa} \|\v x\|_{\m V_{E}^{-1}} \\
        &\quad \leq r(e^r - 1) \alpha 2^s \sqrt{2d \log(1+T/(\kappa \lambda d))} \sqrt{L_{\mu}\kappa} \|\v x\|_{\m V_{E}^{-1}} \tag{\Cref{lem:e_card}} \\
        &\quad \leq r^2 \alpha 2^{s+1} \sqrt{2d \log(1+T/(\kappa \lambda d))} \sqrt{L_{\mu}\kappa} \|\v x\|_{\m V_{E}^{-1}} \tag{$e^r-1\leq 2r$, $\forall r\leq1$} \\
        &\quad \leq \alpha \tau^{(s)} \beta^2 2^{s+1} \sqrt{2d \log(1+T/(\kappa \lambda d))} \sqrt{L_{\mu}\kappa} \|\v x\|_{\m V_{E}^{-1}} \tag{\Cref{lem:small_r}} \\
        &\quad \leq \frac{\alpha}{2} \|\v x\|_{\m V_{E}^{-1}}, 
    \end{align*}
    where the last inequality follows from the definition of $\tau^{(s)}$.
\end{proof}

\subsection{Proof of \Cref{thm:concen}} \label{sec:thm:concen}

Before we start the proof, we introduce one technical lemma:

\begin{lemma} \label{lem:pilot}
    With probability at least $1-\delta/2$, simultaneously for all $s\in[S], t\in[0,T-1]$,
    \begin{align*}
        \norm{\bar{\v\theta}^{(s)}_t - \v\theta^*}_{\m V_{P,t}^{(s)}} \leq \beta, \quad \beta\coloneqq \frac{\kappa}{\sqrt{2}}\sqrt{d\log(1+T/(\kappa\lambda d)) +  \log (2S/\delta)} + B\sqrt{\kappa \lambda}.
    \end{align*}
\end{lemma}

\begin{proof}
    The proof is deferred to \Cref{sec:lem:pilot}.
\end{proof}

\begin{remark}
    In \Cref{lem:pilot}, one could use a fixed-design tail inequality and thereby remove the $\log(T)$ factor in $\beta$. However, our main interest is in reducing the dimension factor $d$, and more importantly in extending this dependence to an effective dimension in future work. Since the results obtained under a fixed-design analysis are tied to the ambient dimension, such an extension would not be possible. For this reason, we instead use a self-normalized inequality for martingale difference sequences, as in \Cref{lem:pilot}.
\end{remark}

Recall the definition of $\cal E_1$ when \Cref{lem:pilot} holds. Then, we can see that \Cref{lem:remainder_final} holds with probability at least $1-\delta/2$.
Now we are ready to start the proof.

\begin{proof} [Proof of \Cref{thm:concen}]
We are going to show that 
\begin{align*}
    \abs{ \v x_{t+1,a}\tp (\what{\v\theta}_{t}^{(s)} - \v\theta^*)} \leq w_{t+1,a}^{(s)}
\end{align*}
holds simultaneously for all $t\in[0,T-1],s\in[S],a\in[K]$ with probability at least $1-\delta$.

Notice that by \Cref{lem:decomp}, we have
\begin{align}
    \abs{\v x_{t+1,a} \tp(\what{\v\theta}_t^{(s)} - \v\theta^*)} \leq \mathrm{(noise)} + \mathrm{(bias)} + \mathrm{(remainder)}. \label{eq:concen1}
\end{align}
First, for the (noise) term, from \Cref{lem:noise}, we replace $\delta\leftarrow \delta/(2TSK)$, and taking union bound gives that, with probability at least $1-\delta/2$, simultaneously for all $t\in[0,T-1],s\in[S],a\in[K]$, 
\begin{align*}
    \mathrm{(noise)} \leq \alpha_1 \|\v x_{t+1,a} \|_{(\m V_{E,t}^{(s)})^{-1}}, \quad \alpha_1 \coloneqq \br{\sqrt{2L_\mu \kappa^2 \log(4TSK/\delta)}  + \frac{\sqrt{\kappa}\log(4TSK/\delta)}{3\sqrt{\lambda}}} ,
\end{align*}
We denote the event as $\cal E_2$ when the above inequality holds.
For the (bias) term, for any $t\in[0,T-1],s\in[S],a\in[K]$,
\begin{align}
    \mathrm{(bias)} &\leq \| \v x_{t+1,a}\|_{(\m H_t^{(s)}(\bar{\v\theta}^{(s)}_t))^{-1}} \cdot \lambda \| \v\theta^*\|_{(\m H_t^{(s)}(\bar{\v\theta}^{(s)}_t))^{-1}} \tag{Cauchy-Schwarz}\\
    &\leq \sqrt{\kappa} \| \v x_{t+1,a}\|_{(\m V_{E,t}^{(s)})^{-1}} \cdot \lambda\|\v\theta^*\|_2 \cdot \lambda_{\min}^{-1/2}(\m H_t^{(s)}(\bar{\v\theta}^{(s)}_t)) \tag{$\m V_{E,t}^{(s)} \preceq \kappa\m H_t^{(s)}(\bar{\v\theta}^{(s)}_t)$} \\
    &\leq B \sqrt{\kappa\lambda} \| \v x_{t+1,a}\|_{(\m V_{E,t}^{(s)})^{-1}} \nonumber \\
    & = \alpha_2 \| \v x_{t+1,a}\|_{(\m V_{E,t}^{(s)})^{-1}}, \nonumber
\end{align}
where the last equality follows from the definition of $\alpha_2$.
For the (remainder) term, from the result of \Cref{lem:remainder_final}, with probability at least $1-\delta/2$ (or on the event $\cal E_1$), simultaneously for all $t\in[0,T-1],s\in[S],a\in[K]$,
\begin{align*}
    \mathrm{(remainder)} \leq \frac{\alpha}{2} \|\v x_{t+1,a}\|_{(\m V^{(s)}_{E,t})^{-1}}.
\end{align*}
Finally, substituting the (noise), (bias), and (remainder) term back to \Cref{eq:concen1}, and taking union bound for $\cal E_1,\cal E_2$, and shifting the index of $t$ by 1 ($t\leftarrow t-1$), we have that, with probability at least $1-\delta$, simultaneously for all $t\in[T], s\in[S], a\in[K]$,
\begin{align*}
    \abs{\v x_{t,a} \tp(\what{\v\theta}_{t-1}^{(s)} - \v\theta^*)} &\leq (\alpha_1 + \alpha_2 + \alpha/2) \|\v x_{t,a}\|_{(\m V^{(s)}_{E,t-1})^{-1}} \\
    &= \alpha \|\v x_{t,a}\|_{(\m V^{(s)}_{E,t-1})^{-1}} \\
    &= w_{t,a}^{(s)},
\end{align*}
completes the proof.
\end{proof}

\subsection{Proof of \Cref{lem:per-round}}

We first introduce the following lemma, showing that the optimal arm is never eliminated until we choose the action at certain level.

\begin{lemma} \label{lem:optimal_arm}
    On the good event $\cal E$, if the action $a_t$ is chosen in level $s_t$, then the optimal arm is never eliminated up to $s_t$, i.e. $a_t^*\in A^{(s)}$ for all $s\in[s_t]$.
\end{lemma}

\begin{proof}
    The proof is deferred to \Cref{sec:lem:optimal_arm}
\end{proof}

Now we start the proof.

\begin{proof}
First, consider the case when $a_t$ is selected by condition (a). 
For the base case $s_t=1$, we can see that $\mu(\v x_{t,a_t^*}\tp \v\theta^*) - \mu(\v x_{t,a_t}\tp\v\theta^*) \leq 8L_\mu2^{-1} =1$.
For the case $s_t>1$, by condition (a) in level $s_t-1$, we have
\begin{align}
    \abs{\v x_{t,a}\tp\v\theta^* - m_{t,a}^{(s_t-1)}} \leq w_{t,a}^{(s_t-1)} \leq 2^{-(s_t-1)}. \label{eq:a1}
\end{align}
Also, by condition (c) in level $s_t-1$, 
\begin{align}
    &m_{t,a}^{(s_t-1)} \geq \max_{a\in A^{(s_t-1)}} m_{t,a}^{(s_t-1)} - 2 \cdot 2^{-(s_t-1)}. \label{eq:c1}
\end{align}
Now, for the per-round regret
\begin{align*}
    \mu(\v x_{t,a_t^*}\tp \v\theta^*) - \mu(\v x_{t,a_t}\tp\v\theta^*) &\leq L_\mu \br{\v x_{t,a_t^*}\tp \v\theta^* - \v x_{t,a_t}\tp\v\theta^*} \tag{\Cref{ass:kappa}, $\v x_{t,a_t^*}\tp \v\theta^* \geq \v x_{t,a_t}\tp\v\theta^*$} \\
    &= L_\mu \sqbr{\uterm{\br{\v x_{t,a_t^*}\tp \v\theta^* - m_{t,a_t^*}^{(s_t-1)}}}{1} + \uterm{\br{m_{t,a_t^*}^{(s_t-1)} - m_{t,a_t}^{(s_t-1)}}}{2} + \uterm{\br{m_{t,a_t}^{(s_t-1)}-\v x_{t,a_t}\tp\v\theta^*}}{3}} 
\end{align*}
For \term 1 and \term 3, by \Cref{eq:a1}
\begin{align*}
    \term 1, \term 3 \leq 2^{-(s_t-1)} = 2 \cdot 2^{-s_t}.
\end{align*}
For \term 2, by \Cref{eq:c1}
\begin{align*}
    \term 2 \leq 2 \cdot 2^{-(s_t-1)} = 4 \cdot 2^{-s_t}
\end{align*}
Substituting \term 1, \term 2, and \term 3 back gives the desired result.

Next, consider the case when $a_t$ is selected by condition (b), which means in level $s_t$,
\begin{align}
    \abs{\v x_{t,a}\tp\v\theta^* - m_{t,a}^{(s_t)}} \leq w_{t,a}^{(s_t)} \leq 1/\sqrt{T}. \label{eq:b1}
\end{align}
Then, for the per-round regret,
\begin{align*}
    \mu(\v x_{t,a_t^*}\tp \v\theta^*) - \mu(\v x_{t,a_t}\tp\v\theta^*) &\leq L_\mu \br{\v x_{t,a_t^*}\tp \v\theta^* - \v x_{t,a_t}\tp\v\theta^*} \\
    &= L_\mu \sqbr{\uterm{\br{\v x_{t,a_t^*}\tp \v\theta^* - m_{t,a_t^*}^{(s_t)}}}{4} + \uterm{\br{m_{t,a_t^*}^{(s_t)} - m_{t,a_t}^{(s_t)}}}{5} + \uterm{\br{m_{t,a_t}^{(s_t)}-\v x_{t,a_t}\tp\v\theta^*}}{6}}.
\end{align*}
For \term 4 and \term 6, by \Cref{eq:b1},
\begin{align*}
    \term 4, \term 6 \leq 1/\sqrt{T}.
\end{align*}
For \term 5, $\term 5\leq 0$ by our choice $a_t$. Substituting \term 4, \term 5, and \term 6 back gives the desired result, completing the proof.
\end{proof}

\subsection{Proof of \Cref{lem:card_P}}

\begin{proof}
    Fix $s\in[S]$. Denote $P^{(s)}(T)$ as $P$ for simplicity. Redefine $P=\{t^{(1)}, \dots, t^{(|P|)}\}$ chronologically, $\m V_0 = \kappa\lambda\I$, $\m V_j = \m V_{j-1} + x_{t^{(j)},a_{t^{(j)}}} x_{t^{(j)},a_{t^{(j)}}} \tp$.
    Then for $t^{(j)}\in P$, by condition (a1),
    \begin{align}
        \norm{\v x_{t^{(j)},a_{t^{(j)}}}}_{\m V_{j-1}^{-1}}^2 \geq \tau^{(s)}. \label{eq:p1}
    \end{align}
    Then,
    \begin{align*}
        |P| \cdot \tau^{(s)} &\leq \sum_{j=1}^{|P|} \min\cbr{1,~\norm{\v x_{t^{(j)},a_{t^{(j)}}}}_{\m V_{j-1}^{-1}}^2} \tag{$\lambda\geq1/\kappa$, $\norm{\v x_{t^{(j)},a_{t^{(j)}}}}_{\m V_{j-1}^{-1}} \leq 1$}\\
        &\leq 2d\log(1 + |P|/(\kappa\lambda d)) \tag{\Cref{lem:yadkori}} \\
        &\leq 2d\log(1 + T/(\kappa\lambda d)),
    \end{align*}
    rearranging $\tau^{(s)}$ to the right-hand side completes the proof.
\end{proof}

\section{Deferred Proof for \Cref{sec:analysis2}}

\subsection{Proof of \Cref{lem:remainder}} \label{sec:lem:remainder}

In this section, we fix $t$ and $s$ and drop the notation for it for simplicity. Also, redefine $r\coloneqq \max_{i\in E}|\v x_{i,a_i}\tp (\bar{\v\theta} - \v\theta^*)|$.

\begin{proof}
In order to control the term $\abs{\dot\mu(\v x_{i,a_i}\tp \check{\v\theta}_i) - \dot\mu(\v x_{i,a_i}\tp\bar{\v\theta})}$, we start by using the fact that $(\log \dot\mu(v))' = \frac{\ddot\mu(v)}{\dot\mu(v)}=1-2\mu(v)$, we have $|(\log \dot\mu(v))'|\leq 1$ and it means that $\log \dot\mu(v)$ is a 1-Lipschitz function.
Therefore, we have
\begin{align*}
    \abs{\log \dot\mu(\v x_{i,a_i}\tp\check{\v\theta}_i) - \log \dot\mu(\v x_{i,a_i}\tp  \bar{\v\theta})} \leq \abs{\v x_{i,a_i}\tp(\check{\v\theta}_i - \bar{\v\theta})},
\end{align*}
where $\check{\v\theta}_i = c_i\bar{\v\theta} + (1-c_i)\v\theta^*$ for some constant $c_i>0$. For the right-hand side
\begin{align*}
    \abs{\v x_{i,a_i}\tp(\check{\v\theta}_i - \bar{\v\theta})} = (1-c_i) \abs{\v x_{i,a_i}\tp(\v\theta^* -\bar{\v\theta})} \leq (1-c_i) r \leq r.
\end{align*}
Substituting this back gives
\begin{align*}
    &\abs{\log \dot\mu(\v x_{i,a_i}\tp\check{\v\theta}_i) - \log \dot\mu(\v x_{i,a_i}\tp \bar{\v\theta})} \leq r \\
    &\quad \Longrightarrow~~ (e^{-r}-1) \dot\mu(\v x_{i,a_i}\tp \bar{\v\theta}) \leq \dot\mu(\v x_{i,a_i}\tp\check{\v\theta}_i) - \dot\mu(\v x_{i,a_i}\tp \bar{\v\theta})\leq (e^r-1) \dot\mu(\v x_{i,a_i}\tp \bar{\v\theta}). 
\end{align*}
Also, since $|e^{-r}-1| \leq |e^r-1|$ for all $r\geq 0$, we have
\begin{align}
    \abs{\dot\mu(\v x_{i,a_i}\tp\check{\v\theta}_i) - \dot\mu(\v x_{i,a_i}\tp \bar{\v\theta})} \leq (e^r-1) \dot\mu(\v x_{i,a_i}\tp \bar{\v\theta}). \label{eq:remainder1}
\end{align}
Now, we begin the proof:
\begin{align*}
    &\abs{\v x\tp \m H^{-1}(\bar{\v\theta}) \br{\sum_{i\in E}\br{\dot\mu(\v x_{i,a_i}\tp \check{\v\theta}_i) - \dot\mu(\v x_{i,a_i}\tp\bar{\v\theta})} \v x_{i,a_i} \v x_{i, a_i}\tp} (\v\theta^* - \bar{\v\theta})} \\
    &= \abs{\sum_{i\in E} \br{\dot\mu(\v x_{i,a_i}\tp \check{\v\theta}_i) - \dot\mu(\v x_{i,a_i}\tp\bar{\v\theta})} \v x\tp \m H^{-1}(\bar{\v\theta})  \v x_{i,a_i} \v x_{i, a_i}\tp (\v\theta^* - \bar{\v\theta})} \\
    &\leq \sum_{i\in E} \abs{\br{\dot\mu(\v x_{i,a_i}\tp \check{\v\theta}_i) - \dot\mu(\v x_{i,a_i}\tp\bar{\v\theta})}} \cdot \abs{\v x\tp \m H^{-1}(\bar{\v\theta})  \v x_{i,a_i}} \cdot \abs{\v x_{i, a_i}\tp (\v\theta^* - \bar{\v\theta})} \\
    &\leq \sum_{i\in E} (e^r -1) \dot\mu(\v x_{i,a_i}\tp \bar{\v\theta}) \cdot \abs{\v x\tp \m H^{-1}(\bar{\v\theta})  \v x_{i,a_i}} \cdot \abs{\v x_{i, a_i}\tp (\v\theta^* - \bar{\v\theta})} \tag{\Cref{eq:remainder1}} \\
    & \leq (e^r -1) \uterm{\sqrt{\sum_{i\in E} \dot\mu(\v x_{i,a_i}\tp \bar{\v\theta}) \br{\v x\tp \m H^{-1}(\bar{\v\theta})  \v x_{i,a_i}}^2}}{1} \uterm{\sqrt{\sum_{i\in E} \dot\mu(\v x_{i,a_i}\tp \bar{\v\theta}) \br{\v x_{i, a_i}\tp (\v\theta^* - \bar{\v\theta})}^2}}{2} . \tag{Cauchy-Schwarz}
\end{align*}
For the \term 1,
\begin{align*}
    \term1 ^2 &= \v x\tp \m H^{-1}(\bar{\v\theta}) \br{\sum_{i\in E} \dot\mu(\v x_{i,a_i}\tp\bar{\v\theta}) \v x_{i,a_i} \v x_{i,a_i}\tp} \m H^{-1}(\bar{\v\theta}) \v x \\
    &\leq \v x\tp \m H^{-1}(\bar{\v\theta}) \m H(\bar{\v\theta}) \m H^{-1}(\bar{\v\theta}) \v x \\
    &\leq \kappa \|\v x\|_{\m V_E^{-1}}^2. \tag{$\m V_E \preceq \kappa \m H(\bar{\v\theta})$}
\end{align*}
For the \term 2,
\begin{align*}
    \term 2 ^2 \leq \sum_{i\in E} L_\mu \br{\v x_{i,a_i}\tp(\v\theta^* - \bar{\v\theta})}^2 \leq |E| r^2 L_\mu.
\end{align*}
Substituting \term 1 and \term 2 back yields the desired result, completing the proof.
\end{proof}

\subsection{Proof of \Cref{lem:e_card}} \label{sec:lem:e_card}

\begin{proof}
In this section, we fix $t$ and $s$ and drop the notation for it for simplicity. Redefine $E = \{t^{(1)}, t^{(2)}, \dots, t^{(|E|)}\}$ chronologically and set $\m V_0 = \kappa \lambda \I$ and $\m V_j = \m V_{j-1} + \v x_{t^{(j)},a_{t^{(j)}}} \v x_{t^{(j)},a_{t^{(j)}}}\tp$.
If $t^{(j)}\in E$, then condition (a) was triggered at level $s$, which means
\begin{align*}
    &w_{t^{(j)},a_{t^{(j)}}}^{(s)} = \alpha \| \v x_{t^{(j)}, a_{t^{(j)}}} \|_{\m V_{j-1}^{-1}} \geq 2^{-s} \\
    &\quad \Longrightarrow~~ \alpha^2 \| \v x_{t^{(j)}, a_{t^{(j)}}} \|_{\m V_{j-1}^{-1}}^2 \geq 4^{-s} \\
    &\quad \Longrightarrow~~\alpha^2\min \cbr{1,~ \| \v x_{t^{(j)}, a_{t^{(j)}}} \|_{\m V_{j-1}^{-1}}^2} \geq 4^{-s},
\end{align*}
where the last inequality holds since $\| \v x_{t^{(j)}, a_{t^{(j)}}} \|_{\m V_{j-1}^{-1}}^2 \leq \|\v x_{t^{(j)}, a_{t^{(j)}}}\|_2^2 / \lambda_{\min}(\m V_{j-1})\leq 1/(\kappa\lambda) \leq 1$ where the last inequality follows from our choice of $\lambda \geq 1/\kappa$.
We can proceed as
\begin{align*}
    \Longrightarrow~~ |E| 4^{-s} &\leq \alpha ^2 \sum_{j=1}^{|E|}\min\cbr{1,~ \| \v x_{t^{(j)}, a_{t^{(j)}}} \|_{\m V_{j-1}^{-1}}^2} \\
    &\leq 2 \alpha^2 d \log (1 + |E|/(\kappa\lambda d)) \tag{\Cref{lem:yadkori}} \\
    &\leq 2 \alpha^2 d \log (1 + T/(\kappa\lambda d)).
\end{align*}
Finally, moving $4^{-s}$ to the right-hand side completes the proof.
\end{proof}

\subsection{Proof of \Cref{lem:pilot}} \label{sec:lem:pilot}

Recall the definition of the loss function
\begin{align*}
    \cal L_t^{(s)}(\v\theta) \coloneqq \sum_{i\in P^{(s)}(t)} \ell (\v x_{i,a_i}, r_i;\v\theta) + \frac{\lambda}{2} \|\v\theta\|_2^2.
\end{align*}
In this section, we fix $t$ and $s$ and drop the notation for it for simplicity. So we simply write as $\cal L(\v\theta) = \sum_{i\in P} \ell(\v x_{i,a_i},r_i;\v\theta) + \frac{\lambda}{2} \|\v\theta\|_2^2$.

\begin{proof}
$\cal L(\v\theta)$ is a strongly convex, differentiable function and $\bar{\v\theta}$ minimizes $\cal L(\v\theta)$ over a convex set $\Theta$. Therefore, we have $\ip{\nabla \cal L(\bar{\v\theta})}{ \v\theta-\bar{\v\theta}} \geq 0$ for all $\v\theta\in\Theta$. Substituting
\begin{align*}
    \nabla \cal L(\bar{\v\theta}) = \sum_{i\in P} \br{\mu(\v x_{i,a_i}\tp \bar{\v\theta}) - r_i} \v x_{i,a_i} + \lambda \bar{\v\theta}, \quad \v\theta \leftarrow \v\theta^*\in\Theta,
\end{align*}
we have
\begin{align*}
    &-\ip{\sum_{i\in P} \br{\mu(\v x_{i,a_i}\tp \bar{\v\theta}) - r_i} \v x_{i,a_i}}{\bar{\v\theta} - \v\theta^*} \geq \lambda \ip{\bar{\v\theta}}{\bar{\v\theta}-\v\theta^*} \\
    &\Longrightarrow~~ -\ip{\sum_{i\in P} \sqbr{\br{\mu(\v x_{i,a_i}\tp \bar{\v\theta}) - \mu(\v x_{i,a_i}\tp \v\theta^*)} - \br{r_i - \mu(\v x_{i,a_i}\tp \v\theta^*)}} \v x_{i,a_i}}{\bar{\v\theta} - \v\theta^*} \geq \lambda \ip{\bar{\v\theta}}{\bar{\v\theta}-\v\theta^*} \\
    &\Longrightarrow~~  \ip{\sum_{i\in P} \eps_i \v x_{i,a_i}}{\bar{\v\theta}-\v\theta^*} \geq \ip{\sum_{i\in P} \br{\mu(\v x_{i,a_i}\tp \bar{\v\theta}) - \mu(\v x_{i,a_i}\tp \v\theta^*)} \v x_{i,a_i}}{\bar{\v\theta}-\v\theta^*} + \lambda \ip{\bar{\v\theta}}{\bar{\v\theta}-\v\theta^*}. \\
    &\Longrightarrow~~  \uterm{\ip{\sum_{i\in P} \eps_i \v x_{i,a_i}}{\bar{\v\theta}-\v\theta^*}}{1} \geq \uterm{\ip{\sum_{i\in P} \br{\mu(\v x_{i,a_i}\tp \bar{\v\theta}) - \mu(\v x_{i,a_i}\tp \v\theta^*)} \v x_{i,a_i} + \lambda(\bar{\v\theta} - \v\theta^*)}{\bar{\v\theta}-\v\theta^*}}{2} + \uterm{\lambda \ip{\v\theta^*}{\bar{\v\theta}-\v\theta^*}}{3}.
\end{align*}
For the \term 1,
\begin{align*}
    \term 1 \leq \norm{\sum_{i\in P} \eps_i \v x_{i,a_i}}_{\m V_P^{-1}} \cdot \|\bar{\v\theta} - \v\theta^*\|_{\m V_P} \tag{Cauchy-Schwarz}.
\end{align*}
For the \term 2,
\begin{align*}
    \term 2 &= \sum_{i\in P} (\bar{\v\theta} -\v\theta^*)\tp \dot\mu(\v x_{i,a_i}\tp \check{\v\theta}_i) \v x_{i,a_i} \v x_{i, a_i}\tp(\bar{\v\theta} -\v\theta^*) + (\bar{\v\theta} -\v\theta^*)\tp \lambda \I (\bar{\v\theta} -\v\theta^*) \tag{mean-value}\\
    &= (\bar{\v\theta} -\v\theta^*)\tp \br{\sum_{i\in P} \dot\mu(\v x_{i,a_i}\tp \check{\v\theta}_i) \v x_{i,a_i} \v x_{i, a_i}\tp  + \lambda \I} (\bar{\v\theta} -\v\theta^*) \\
    &\geq \frac{1}{\kappa} (\bar{\v\theta} -\v\theta^*)\tp \br{\sum_{i\in P}  \v x_{i,a_i} \v x_{i, a_i}\tp  +  \kappa \lambda \I} (\bar{\v\theta} -\v\theta^*) \tag{$\check{\v\theta}_i\in\Theta$}\\
    &\geq \frac{1}{\kappa} \|\bar{\v\theta} -\v\theta^*\|_{\m V_P}^2.
\end{align*}
For the \term 3,
\begin{align*}
    \term 3 \geq - \lambda \|\v\theta^*\|_{\m V_P^{-1}} \cdot \|\bar{\v\theta} - \v\theta^*\|_{\m V_P} \geq -B \sqrt{\frac{\lambda}{\kappa}}   \|\bar{\v\theta} - \v\theta^*\|_{\m V_P},
\end{align*}
where the first inequality follows from the Cauchy-Schwarz inequality, and the second inequality holds since $ \|\v\theta^*\|_{\m V_P^{-1}}\leq \|\v\theta^*\|_2/\lambda_{\min}^{1/2}(\m V_P) \leq B/\sqrt{\kappa\lambda}$.

Substituting \term 1, \term 2, and \term 3 back, and dividing both sides with $\frac{1}{\kappa}\|\bar{\v\theta} - \v\theta^*\|_{\m V_P}$ gives
\begin{align*}
    \|\bar{\v\theta} - \v\theta^*\|_{\m V_P} \leq \kappa \uterm{{\norm{\sum_{i\in P} \eps_i \v x_{i,a_i}}_{\m V_P^{-1}}}}{4} + B \sqrt{\kappa\lambda}.
\end{align*}
For \term 4, we prepare to apply the self-normalized bound for vector-valued martingales, which is the well known result from Theorem 1 of \citet{abbasi2011improved}.

Redefine $P = \{t^{(1)}, t^{(2)}, \dots,t^{(|P|)}\}$ chronologically, and let $\m V_0=\kappa \lambda \I$ and $\m V_j = \m V_{j-1} + \v x_{t^{(j)}, a_{t^{(j)}}} \v x_{t^{(j)}, a_{t^{(j)}}}\tp$. Then
\begin{align*}
    A_4 = {\norm{\sum_{j=1}^{|P|} \eps_{t^{(j)}} \v x_{{t^{(j)}},a_{t^{(j)}}}}_{\m V_{|P|}^{-1}}}.
\end{align*}
We can see that $\v x_{{t^{(j)}},a_{t^{(j)}}}$ is $\cal F_{t^{(j)}}$-measurable, and $\eps_{t^{(j)}}$ is $\cal F_{t^{(j+1)}}$-measurable. Also, $\E[\eps_{t^{(j)}}\mid \cal F_{t^(j)}]=0$, $|\eps_{t^{(j)}}|\leq1$, therefore $1/2$-subGaussian noise. Applying \Cref{lem:selfnorm}, simultaneously for all $j\geq 0$, with probability at least $1-\delta$,
\begin{align*}
    \term 4 \leq \frac{1}{\sqrt{2}}\sqrt{d\log(1+|P|/(\kappa\lambda d)) +  \log (1/\delta)} \leq \frac{1}{\sqrt{2}}\sqrt{d\log(1+T/(\kappa\lambda d)) +  \log (1/\delta)}.
\end{align*}
Finally, substituting \term 4 back, replacing $\delta \leftarrow \delta/(2S)$, and taking the union bound for all $s\in[S]$ yields that, with probability at least $1-\delta/2$, simultaneously for all $t\in[T], s\in[S]$, the desired result holds. 
\end{proof}

\subsection{Proof of \Cref{lem:optimal_arm}} \label{sec:lem:optimal_arm}

Denote the good event $\cal E$ when \Cref{thm:concen} holds.

\begin{proof}
We prove this by induction.

First, for $s=1$, $a_t^* \in A^{(1)}=[K]$.
Set $s'\in[s_t]$ and assume that for $s=s'$, $a_t^*\in A^{(s')}$.

Now, Consider $s=s'+1$. On the good event $\cal E$,
\begin{align*}
    \abs{m_{t,a^*}^{(s')}- \v x_{t,a^*}^\top \v\theta^*} \leq w_{t,a^*}^{(s')} \leq 2^{-s'},
\end{align*}
where the first inequality follows from \Cref{thm:concen}, and the second inequality holds since the algorithm reach level $s'+1$ implies that the condition (a) in level $s'$ is not satisfied. Denote $\hat a= \max_{a\in A^{(s')}} m_{t,a}^{(s')}$. Then,
\begin{align*}
    &m_{t,a^*}^{(s')} \geq \v x_{t,a^*}^\top \v\theta^* - 2^{-s'} \geq \v x_{t,\hat a}^\top \v\theta^* - 2^{-s'} \geq m^{(s')}_{t,\hat a} - 2 \cdot 2^{-s'} \\
    &\quad\Longrightarrow~~ m_{t,a^*}^{(s')} \geq \max_{a\in A^{(s')}} m_{t,a}^{(s')} - 2 \cdot 2^{-s'},
\end{align*}
which implies that $a^*\in A^{(s'+1)}$ by condition (c) in level $s'$, completing the proof.
\end{proof}

\section{Auxiliary Lemmas}

\begin{lemma} [Bernstein inequality] \label{lem:bernstein}
    For independent random variables $X_1,\dots,X_n$ with $\E[X_i]=0$, $|X_i|\leq b$ almost surely, and $\sum_{i=1}^n \Var(X_i)\leq V$, then for all $L\geq 0$,
    \begin{align*}
        \P\br{\abs{\sum_{i=1}^n X_i} \geq \sqrt{2VL} + \frac{bL}{3}} \leq 2e^{-L}.
    \end{align*}
\end{lemma}

\begin{lemma} [Elliptical potential lemma, Lemma 11 of \citet{abbasi2011improved}] \label{lem:yadkori}
    For any $\lambda > 0$ and sequence $\{\v x_t\}_{t=1}^T \in \mathbb{R}^d$, define $\m V_0 =\lambda \I$ and  $\m V_t = \m V_{t-1} + \v x_i \v x_i\tp$. Then, provided that $\|\v x_t\|_2 \leq L$ holds for all $t\in[T]$, we have
    \begin{align*}
        \sum_{t=1}^T \min \cbr{1,~\|\v x_t\|_{\m V_{t-1}^{-1}}^2} \leq 2 \log \frac{\det \m V_T}{\det \m V_0} \leq 2d\log \br{1 + \frac{TL^2}{d\lambda}}.
    \end{align*} 
\end{lemma}

\begin{lemma} [Theorem 1 of \citet{abbasi2011improved}] \label{lem:selfnorm}
    Let $\{\cal F_t\}_{t=1}^\infty$ be a filtration. Let $\{\v x_t\}_{t=1}^\infty$, $\{\eps_t\}_{t=1}^\infty$ be stochastic processes such that $\v x_t\in\R^d$ is $\cal F_t$-measurable with $\|\v x_t\|_2\leq L$, and $\eps_t\in\R$ is $\cal F_{t+1}$-measurable and $R$-subGaussian condition on $\cal F_t$. Define $\m V_0=\lambda \I$ and $\m V_t = \lambda \I + \sum_{i=1}^t \v x_i \v x_i\tp$. Fix $\delta\in(0,1]$. Then with probability at least $1-\delta$, simultaneously for all $t\geq 0$, 
    \begin{align*}
        \norm{\sum_{i=1}^t \eps_i \v x_i}_{\m V_t^{-1}}^2 \leq 2 R^2 \br{\log \frac{\det \m V_t}{\det \m V_0} + \log(1/\delta)} \leq 2 R^2 \br{d\log\br{1+\frac{tL^2}{d\lambda}} + \log(1/\delta)}.
    \end{align*}
\end{lemma}